\PassOptionsToPackage{table}{xcolor}
\documentclass[11pt,a4paper]{article}
\usepackage{technical_report}

\usepackage{amsmath,amsfonts,bm}

\def\eqref#1{equation~\ref{#1}}

\def\1{\bm{1}}

\DeclareMathAlphabet{\mathsfit}{\encodingdefault}{\sfdefault}{m}{sl}
\SetMathAlphabet{\mathsfit}{bold}{\encodingdefault}{\sfdefault}{bx}{n}

\newcommand{\R}{\mathbb{R}}

\usepackage{hyperref}
\hypersetup{colorlinks=true,allcolors=reportaccent,bookmarksnumbered=true}
\usepackage{url}
\usepackage{graphicx}
\usepackage{wrapfig}
\usepackage{array}
\usepackage{tabularx}
\usepackage{multirow}
\usepackage{booktabs}
\usepackage{capt-of}
\usepackage{placeins}
\usepackage{flafter}
\usepackage{algorithm}
\usepackage{algpseudocode}
\usepackage{subcaption}
\usepackage{threeparttable}

\newcommand{\authsep}{{\ }}

\newcommand{\ReportTableStyle}{\small\setlength{\tabcolsep}{4pt}\renewcommand{\arraystretch}{1.1}}

\newcommand{\draftmode}{0}
\newcommand{\todo}[1]{\ifnum\draftmode=1\relax\textcolor{red}{[\textbf{TODO:} #1]}\fi}
\newcommand{\yaobo}[1]{\ifnum\draftmode=1\relax\textcolor{blue}{#1}\fi}

\newcolumntype{Y}{>{\centering\arraybackslash}X}
\definecolor{highlight}{RGB}{255,215,0}

\usepackage{adjustbox}
\usepackage{colortbl}
\usepackage{needspace}
\usepackage[capitalize]{cleveref}
\crefname{section}{Sec.}{Secs.}
\Crefname{section}{Section}{Sections}
\crefname{table}{Tab.}{Tabs.}
\Crefname{table}{Table}{Tables}
\AddToHook{cmd/section/before}{\Needspace{6\baselineskip}}

\providecommand{\R}{\mathbb{R}}

\hypersetup{pdftitle={Pretraining Simple Vision-Language-Action Models for Strong and Generalizable Robotic Manipulation},pdfauthor={}}

\title{\vspace{10pt}CogACT2: Pretraining Simple Vision-Language-Action Models for Strong and Generalizable Robotic Manipulation}

\author{%
	Yaobo Liang\textsuperscript{1$\star$},\authsep
	Lin Luo\textsuperscript{1$\star$},\authsep
	Chengtang Yao\textsuperscript{1$\star$},\authsep
    Zhenhua Liu\textsuperscript{1$\star$},\authsep
	Xi Chen\textsuperscript{1$\star$},\authsep
	Sicheng Xu\textsuperscript{1$\star$},\authsep
	Xingyuming Liu\textsuperscript{2$\diamond$},\authsep
	Qingwen Yang\textsuperscript{2$\diamond$},\authsep
	Huizhi Liang\textsuperscript{3$\diamond$},\authsep
	Lingyu Kong\textsuperscript{3$\diamond$},\authsep
	Zili Lin\textsuperscript{4$\diamond$},\authsep
	Huanshuo Dong\textsuperscript{4$\diamond$},\authsep
	Zhiyan Zhang\textsuperscript{4$\diamond$},\authsep
	Qixiu Li\textsuperscript{3$\diamond$},\authsep
	Zhiyuan Feng\textsuperscript{3$\diamond$},\authsep
	Yuchen Lin\textsuperscript{2$\diamond$},\authsep
	Yizhong Zhang\textsuperscript{1},\authsep
	Hao Chen\textsuperscript{1},\authsep
	Lijun Sun\textsuperscript{1},\authsep
	Chenren Xu\textsuperscript{2},\authsep
	Ziqing Gu\textsuperscript{3},\authsep
	Xin Jin\textsuperscript{4},\authsep
	Baining Guo\textsuperscript{1},\authsep
	Jiaolong Yang\textsuperscript{1$\dag$}\\[5pt]
	{\small\color{reportmuted}\textsuperscript{1}Microsoft Research Asia \ 
	\small\color{reportmuted}\textsuperscript{2}Peking University \ 
	\small\color{reportmuted}\textsuperscript{3}Tsinghua University \  \small\color{reportmuted}\textsuperscript{4}Eastern Institute of Tech., Ningbo\\
	\small\color{reportmuted}\textsuperscript{$\star$}Core Contributor \quad \small\color{reportmuted}\textsuperscript{$\diamond$}Intern at MSRA \quad
	\small\color{reportmuted}\textsuperscript{$\dag$}Project Lead}
}

\date{}

\begin{document}
\maketitle
\thispagestyle{empty}
\noindent\begin{minipage}{\textwidth}
    \centering
    \captionsetup{hypcap=false}
\includegraphics[width=0.999\linewidth]{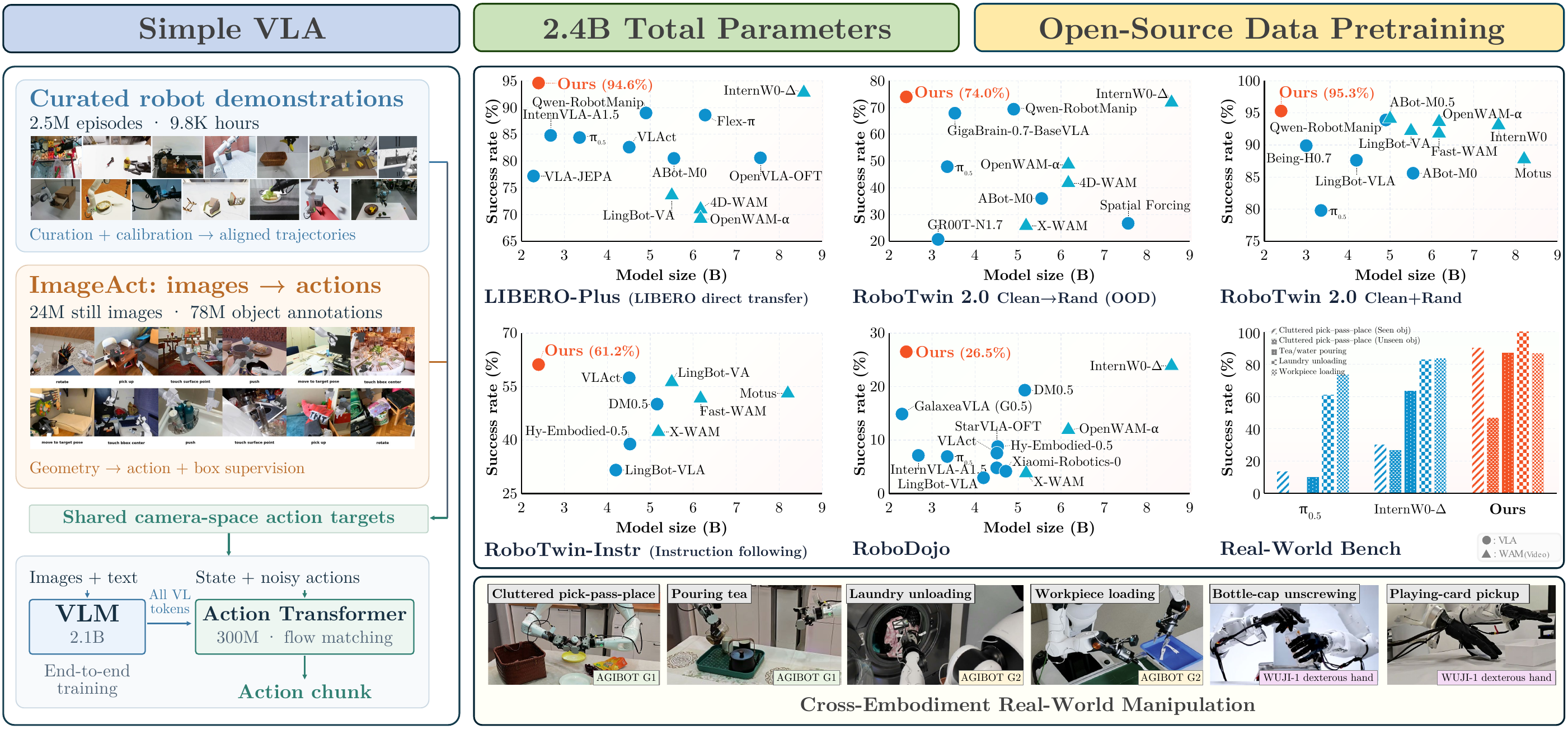}
\vspace{-18pt}
    \captionof{figure}{
        Our simple and compact VLA model achieves strong performance on diverse  manipulation benchmarks and  real-world tasks across different robot embodiments. It is pretrained on data from public sources (none tailored to any target benchmark or real-world evaluation task).
        }
    \label{fig:teaser}
\end{minipage}
\par

\begin{abstract}

\vspace{-4pt}
Since the emergence of vision-language-action (VLA) models, research in generalist robot manipulation has increasingly focused on more complex models, incorporating diverse priors and broader modeling frameworks. While they have delivered increasingly strong results, we investigate how far a simple, general-purpose VLA architecture can go through effective supervision and pretraining.

\vspace{3.5pt}

We present CogACT2, a deliberately simple VLA model for generalist robot manipulation. Our model combines a vision-language model with an action transformer and contains only 2.4B total parameters. We curate heterogeneous robot demonstrations and unify their actions in camera space through robust calibration. To extend supervision, we introduce ImageAct, which uses scene geometry to convert still images into robot motion targets. Despite its simplicity and compact size, our model outperforms a broad range of state-of-the-art models on diverse simulation benchmarks and achieves strong performance across three real robots on bimanual manipulation and dexterous hand tasks.
Comprehensive analysis demonstrates the effectiveness of ImageAct and identifies key training and inference choices that contribute to strong VLA performance.
These results demonstrate that simple VLA architectures can achieve strong generalist manipulation capabilities through scalable supervision and effective pretraining.

\vspace{3.5pt}

\textbf{Models:} \url{https://huggingface.co/collections/microsoft/cogact2}

\textbf{Code:} \url{https://github.com/microsoft/cogact2}

\textbf{Project:} \url{https://microsoft.github.io/CogACT2}

\vspace{-4pt}

\end{abstract}
\section{Introduction}
\label{sec:intro}

Generalist robot manipulation requires policies that translate visual observations and language instructions into actions across diverse tasks, environments, and embodiments. Vision-language-action (VLA) models approach this goal by adapting vision-language models (VLMs) to robot control~\cite{brohan2023rt2,kim2024openvla,black2024pi0,li2024cogact,bjorck2025groot}. Recent work has explored increasingly elaborate modeling approaches, incorporating explicit reasoning and hierarchical structures~\cite{zawalski2024ecot,belkhale2024rth}, geometric priors~\cite{qu2025spatialvla,li2026spatial,han2026geometric}, and broader frameworks such as world action models~\cite{ye2026world,li2026lingbotva,luo2026beingh07}.
Alongside these advances, the principle of parsimony and the emphasis on learning from data and computation~\cite{sutton2019bitter} motivate us to explore \emph{how much can be achieved with a simple, general-purpose VLA architecture through improvements in supervision and pretraining}.

We investigate this question with \textbf{CogACT2}, a deliberately simple VLA model for language-conditioned robot manipulation. 
Following the design principle of coupling pretrained vision-language understanding with a dedicated action-generation module, as exemplified by CogACT~\cite{li2024cogact} and other early models such as $\pi_0$~\cite{black2024pi0}, our model pairs a pretrained VLM with an action transformer. Focusing on diverse pretraining supervision, we curate heterogeneous robot demonstrations from existing datasets and unify their action spaces through systematic data processing and robust calibration. We also incorporate supplementary egocentric human action data to broaden scene and action coverage and support dexterous hand control.

To extend action supervision beyond recorded trajectories, we introduce \textbf{ImageAct}, which converts \emph{still images} into instruction-conditioned robot motion targets. Still-image collections contain objects, viewpoints, and spatial arrangements for which robot demonstrations may be unavailable. ImageAct uses scene geometry to construct motion targets associated with language instructions, turning these visual examples into action-learning samples. The same scene can support different target motions under different instructions. This supervision is geometric rather than evidence of successful physical execution: it complements recorded interaction without replacing the contact and dynamics information present in robot demonstrations. Our experiments show that pretraining with ImageAct data significantly improves the downstream performance of our VLA model.

We systematically evaluate our pretrained model across a wide range of simulation benchmarks and real-world tasks.
Despite its simple architecture and compact 2.4B size, our model outperforms a broad range of state-of-the-art models across simulation benchmarks. For example, it achieves 94.6\% on LIBERO-Plus~\cite{liberoplus2025}, and 74.0\% on RoboTwin 2.0~\cite{chen2025robotwin2} under clean-to-random transfer and 95.3\% under its mixed clean/randomized training setting. Real-world experiments further demonstrate strong performance across diverse tasks on three robots, including bimanual manipulation and dexterous hand control.

Our analysis uncovers insights into training and inference choices that contribute to strong VLA performance.
Strengthening language guidance through language dropout during training and classifier-free guidance~\cite{ho2022classifierfree} at inference consistently improves success rates and generalization.
Consistent camera-space supervision, enabled by recovering camera extrinsics for hard-to-calibrate data, outperforms mixed-space training that retains robot-space actions. Auxiliary target-object bounding-box prediction by the action transformer during pretraining also brings gains in grounding and generalization.

\textbf{Our contributions} can be summarized as follows:
\begin{itemize}
    \item We demonstrate that effective supervision and pretraining enable a simple, compact VLA architecture to achieve strong manipulation performance across diverse benchmarks and real-world tasks.
    \item We develop a scalable action-supervision pipeline using diverse training data and introduce ImageAct, a new way to expand supervision by converting still images into instruction-conditioned robot motion targets.
    \item We uncover insights into the factors that drive strong VLA performance such as language guidance, action-space consistency, and auxiliary grounding supervision.
\end{itemize}

{We will release our models and code for training, evaluation, and data processing} to provide a compact, capable, and easy-to-adapt foundation for further research. We hope this work inspires further progress toward strong and general-purpose policies that are built on simple architectures and scalable supervision.

\section{Related Work}
\label{sec:related}
\paragraph{Vision-language-action models.}
VLA models adapt pretrained VLMs to robot control, differing mainly in how actions are predicted. Some methods predict discretized actions~\cite{brohan2023rt2,kim2024openvla}, while others combine VLMs with continuous flow- or diffusion-based generators~\cite{black2024pi0,li2024cogact}. Recent works further explore alternative action prediction, including parallel regression \cite{kim2025openvlaoft} for efficient continuous action prediction and soft-prompted transformers \cite{zheng2025xvla} for flexible cross-embodiment adaptation. Another line of work freezes the pretrained VLM and employs a separate flow-matching action expert for continuous action prediction \cite{shukor2025smolvla}. Beyond action prediction, other approaches incorporate skill grounding~\cite{ahn2022saycan}, reasoning \cite{zawalski2024ecot}, or geometric representations \cite{qu2025spatialvla} to enhance VLA capabilities. We retain the VLM-action-generator design and train it end-to-end, with the main emphasis on the supervision that a simple architecture can absorb.

\paragraph{World action models.}
Some approaches adapt video generators to robot control~\cite{shen2025videovla,kim2026cosmospolicy}, while others connect actions to predictive visual or latent dynamics~\cite{ye2026world,li2026lingbotva,luo2026beingh07}. To avoid the overhead of video rollout at deployment, \cite{internw0} combines predictive and geometric supervision with pretrained visual and video representations to generate actions directly. We use a pretrained VLM and direct action generation, without a video-generation backbone or auxiliary future-observation prediction.

\paragraph{Scaling pretraining.}
Recent work scales learning through heterogeneous data sources \cite{openx2024,octo2024,black2025pi05,yuan2026qwenrobotmanip}. Visual pretraining supports downstream control~\cite{nair2022r3m,xiao2022mvp,majumdar2023vc1}, while other work explores continuous representation pretraining for VLA transfer~\cite{vlact2026}. Human videos provide a scalable and low-cost source of motion data, which can be learned by latent action extraction~\cite{ye2025lapa} or reconstructed hand motion~\cite{li2026scalable}. Spatial and 3D data provide supervision for spatial reasoning, affordance grounding, and geometric understanding~\cite{chen2024spatialvlm,yuan2024robopoint,hong2023threedllm}. Our data pipeline brings robot motion, supplementary human trajectories, and image-derived motion targets into one action interface, emphasizing curation and alignment alongside scale. Our image-derived supervision supplies explicit motion targets from still images while retaining the distinction between constructed geometry and executed interaction.

\section{Model}
\label{sec:model}

\begin{figure}[t!]
\centering
\includegraphics[width=1\linewidth]{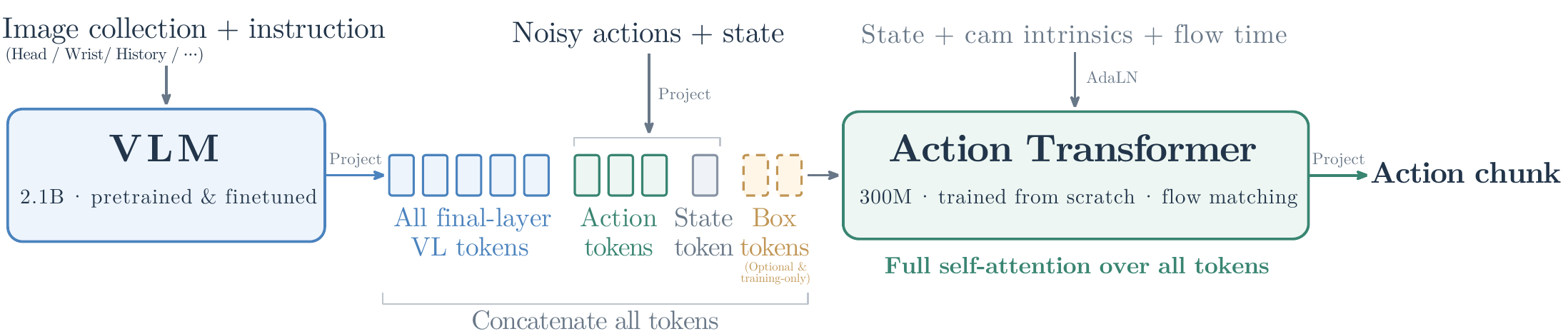}

\caption{\textbf{Model architecture.} A pretrained VLM is paired with an action transformer. All final-layer vision-language tokens are jointly processed with action and state tokens through full self-attention, with state, camera, and flow-time conditioning.
}
\label{fig:architecture}
\end{figure}

\Cref{fig:architecture} presents our overall architecture: the VLM encodes images and language, and the action transformer maps the resulting token representations to actions.

\subsection{Vision-language and action processing}
\paragraph{Vision-language modeling.}
At observation time $t$, the policy receives a collection of image observations $\mathcal I_t$ and a language instruction $\ell$. %
We use RynnBrain1.1-2B~\cite{li2026rynnbrain11} as the pretrained VLM $E_\phi$ to jointly encode the images and instruction. Its complete {final-layer} token sequence is
\begin{equation}
    \mathbf H^{\mathrm{VL}}=E_\phi(\mathcal I_t,\ell)\in\R^{S\times d},
    \label{eq:vlm}
\end{equation}
where $S$ is the sequence length and $d$ the VLM hidden width. A learned projection $P_\theta$ maps each token to the action transformer width. We preserve {all visual and language tokens}, giving the action model access to token-level context without pooling or learned summary queries.

\paragraph{Image observations.}
The default image collection input $\mathcal{I}_t$ to our model includes four images: one current frame from the main camera, two current frames from the left and right wrist cameras, and one history frame at time $t-\delta_t$ from the main camera where we set $\delta_t=1$\,s. If the wrist camera frames are unavailable, we simply omit them. One can flexibly adjust the input images during finetuning such as adding anchor frames or wrist camera history frames when necessary.

\paragraph{Action transformer.}
For an action horizon $T$, each action vector is embedded as one token. At flow time $\tau$, the action transformer input is
\begin{equation}
\mathbf Z_\tau=\operatorname{Concat}
[P_\theta(\mathbf H^{\mathrm{VL}}),\mathbf X^a_\tau,\mathbf x^s,\mathbf X^b],
\label{eq:joint}
\end{equation}
where $\mathbf X^a_\tau$ contains noisy action tokens, $\mathbf x^s$ is a state token, and $\mathbf X^b$ contains $B$ optional box tokens. The action transformer uses \emph{full self-attention}: all retained token groups can attend to one another at every layer. The projected vision-language tokens are themselves updated by these layers, rather than serving only as conditioning features.

\paragraph{State, camera, and flow-time conditioning.}
State, camera intrinsics $\boldsymbol{\kappa}$ represented by the field-of-view (FOV) angles of the input cameras, and flow time condition every action transformer layer through shared AdaLN~\cite{chen2024pixart}. The state is also embedded as a token that participates in self-attention.

\subsection{A unified action and state interface}
\label{sec:interface}
\label{subsec:robot-interface}
\label{app:action-state}

We represent motion using \emph{camera-space end-effector (EEF) poses} in the reference camera frame (typically the head camera), with EEF coordinate conventions standardized across platforms. The representation includes translation, rotation, and either gripper commands or articulated-hand poses~\cite{li2026scalable,romero2017mano}. We \emph{do not include arm joint angles} because joint configurations vary substantially across embodiments.

\paragraph{Action representation.} Our action representation can be written as $\mathbf{a}_t = [\mathbf{a}^{\mathrm{grip}}_t; \mathbf{a}^{\mathrm{hand}}_t]$, which contains gripper action and articulated hand action. For each gripper arm, the action specifies an EEF translation, a 6D rotation represented by the first two columns of the rotation matrix \cite{zhou2019continuity}, and a scalar gripper opening. The left and right arms together occupy $(3+6+1)\times2=20$ dimensions. Following \cite{li2026scalable}, we design the action for each hand as the wrist EEF translation and 6D rotation, together with three Euler angles for each of 15 finger joints as defined in the MANO hand model~\cite{romero2017mano}. The two hands therefore occupy $(3+6+15\times3)\times2=108$ dimensions.

\paragraph{State and availability masks.}
The state uses the same 128-dimensional layout and appends 10 MANO shape parameters $\boldsymbol{\beta}$ for each hand: $
    \mathbf{s} =
    [\mathbf{s}^{\mathrm{motion}};
     \boldsymbol{\beta}^{\mathrm{L}};
     \boldsymbol{\beta}^{\mathrm{R}}]$.
Every action and state vector is paired with an equally-sized binary mask indicating whether each dimension is available for that sample: $\mathbf{m}^{a}_t\in\{0,1\}^{128}$ and $\mathbf{m}^{s}\in\{0,1\}^{148}$.

\begin{table}[t!]
\centering
\caption{\textbf{Retained robot demonstrations for pretraining after processing.} The source
inventory spans real and simulated single-arm and bimanual platforms.
}
\label{tab:robot-data}
\small
\setlength{\tabcolsep}{3pt}
\begin{tabularx}{\textwidth}{@{}lXcrr@{}}
\toprule
Source & Embodiment & \!Environment\! & Episodes & Hours \\
\midrule
AgiBotWorldBeta~\cite{bu2025agibot}\! & AGIBOT G1 & Real & 946,507 & 2,376 \\
RoboChallenge~\cite{yakefu2025robochallenge} & ALOHA, ARX5, Franka Panda, UR5 & Real & 12,460 & 173 \\
RoboCOIN~\cite{wu2025robocoin} & AgileX Cobot Magic and Decoupled Magic & Real & 24,436 & 145 \\
RoboMIND~\cite{wu2024robomind} & AgileX, Franka Panda, dual Franka FR3 & Real/Sim & 6,166 & 11 \\
RoboMIND2.0~\cite{hou2025robomind2} & AgileX & Real/Sim & 4,030 & 176 \\
InternData-A1~\cite{tian2025interndataa1} & Franka Panda, AGIBOT G1, ARX Lift-2, AgileX Split ALOHA & Sim & 451,089 & 3,131 \\
InternData-M1~\cite{contributors2025interndatam1} & AgileX, Franka Panda (with Robotiq gripper) & Sim & 458,788 & 1,618 \\
RoboCasa365~\cite{nasiriany2026robocasa365} & Franka Panda & Sim & 592,976 & 2,215 \\
\midrule
Total & & & 2,496,452 & 9,844 \\
\bottomrule
\end{tabularx}
\end{table}

\subsection{Flow-matching action generation}
\label{sec:flow-matching}
We generate an action chunk $\mathbf A = [\mathbf a_t, \ldots, \mathbf a_{t+T-1}]$ with a default chunk size of $T=30$ through flow matching~\cite{lipman2022flowmatching}, using an $x_0$-prediction parameterization. Given a noisy chunk $\mathbf A_\tau=(1-\tau)\mathbf A_0+\tau\epsilon$, where $\mathbf A_0$ is the clean action chunk, $\tau$ is the flow time, and $\epsilon$ is noise, the action transformer directly predicts the clean chunk:
\begin{equation}
\hat{\mathbf A}_0=f_\theta(\mathbf A_\tau,\tau;
\mathbf H^{\mathrm{VL}},\mathbf s,\mathbf m^s,\boldsymbol{\kappa},\mathbf M^a).
\label{eq:action-prediction}
\end{equation}
Here, $\mathbf M^a$ stacks per-step action masks. Predictions are read from the action-token positions of the joint sequence in \cref{eq:joint}. Training supervises available numerical targets and propagates gradients through the projection to the VLM. At inference, we iteratively refine a noisy action chunk using the model's clean-action predictions.

\paragraph{Optional box generation.}
The model can additionally predict target-object 2D and 3D boxes through optional box tokens that participate in self-attention. Box supervision is provided by ImageAct samples (Sec.~\ref{sec:image2action}); no box prediction is performed for other training data, or during inference.

\FloatBarrier
\section{Scaling Diverse Action Supervision}
\label{sec:data}

We scale action supervision for pretraining through three sources: robot demonstrations ground learning in physical interaction, ImageAct synthesizes action targets from still images to broaden visual and spatial coverage, and human videos provide hand motion from everyday activities.

\subsection{Robot Data}
\label{subsec:robot-data}
\label{app:robot-data}

We gather real-world and simulated robot demonstrations across diverse embodiments from AgiBotWorldBeta~\cite{bu2025agibot}, RoboChallenge~\cite{yakefu2025robochallenge},
RoboCOIN~\cite{wu2025robocoin}, RoboMIND~\cite{wu2024robomind},
RoboMIND2.0~\cite{hou2025robomind2}, InternData-A1~\cite{tian2025interndataa1},
InternData-M1~\cite{contributors2025interndatam1}, and
RoboCasa365~\cite{nasiriany2026robocasa365}.
To improve data quality and reduce noise, we develop a comprehensive data curation and processing pipeline that corrects video metadata and invalid instructions, detects corrupted frames, recovers missing states and actions, and standardizes coordinate conventions. We also detect and filter abrupt motion, temporal discontinuities, and static redundancy.
The resulting corpus contains about \textbf{9.8K hours} of demonstrations spanning 2.5M episodes. Table~\ref{tab:robot-data} lists the retained episodes and hours for each source and embodiment.

\begin{figure}[t!]
\centering
\includegraphics[width=1\linewidth]{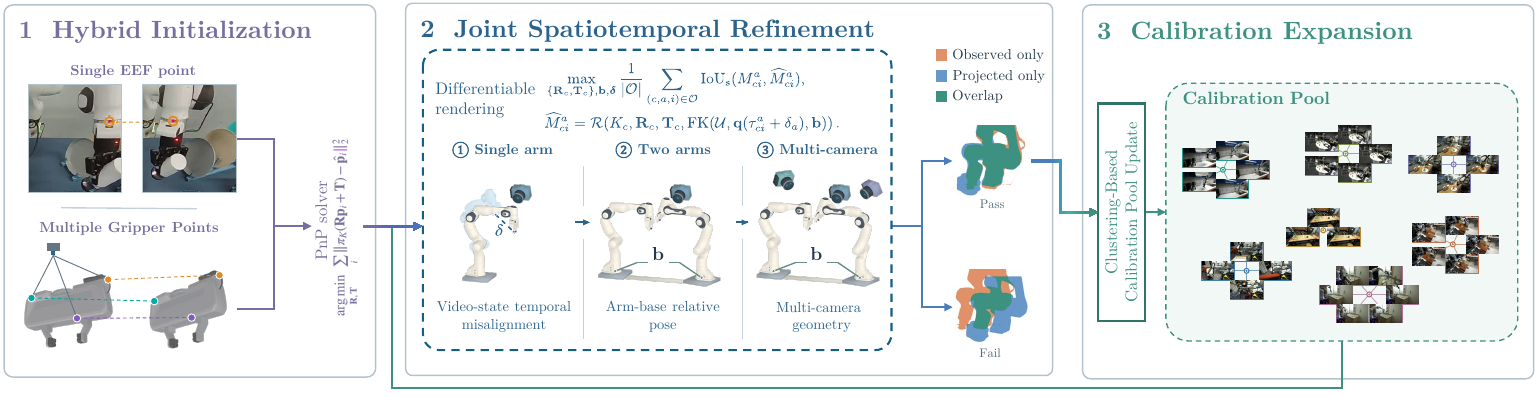}
\caption{\textbf{Camera calibration pipeline.} A three-stage camera-robot pose estimation approach that addresses limited EEF visibility, video--state misalignment, multi-camera configurations, and other robustness challenges in processing robot data.
}
\label{fig:cam_calib}
\end{figure}

\paragraph{Data filtering and standardization.}
We first correct mislabeled camera names and detect corrupted frames according to the intensity statistics.
We also inspect instruction data and correct or remove incomplete, corrupted, or ambiguous text. 

For motion data, we first recover the missing or incomplete states and actions by forward kinematics from joint positions. We then standardize the coordinate conventions for the robot base and the tool center point to ensure spatial consistency across the training data.
Beyond coordinate inconsistency, robot motion data often contain redundant or irregular motions that provide unhelpful or even harmful supervision. We repair the data by excluding invalid intervals or entire episodes based on four detection criteria: (1) Abrupt motion changes are identified by checking whether end-effector translation speed and rotation change exceed thresholds. (2) Temporal discontinuities are detected by identifying consecutive timestamp gaps that reach $1\text{s}$. (3) Redundant static motion is identified through two steps: we detect static frames by velocity limits and multi-scale image differences, while safeguarding key events; we then consolidate these frame-level detections into continuous intervals. (4) Redundant return/reset motion is truncated by verifying round-trip displacements toward the initial position, which locates late-stage turning points.

\subsection{Camera Calibration}
\label{sec:camera-calibration}
Unifying actions in camera space requires reliable camera parameters, which are unavailable in many datasets. This consistency matters: mixing camera-space and robot-space action supervision lowers the downstream finetuning performance in our ablation (\cref{tab:ablation-all}, Sec.~\ref{sec:analysis}).
Prior calibration methods~\cite{xie2025caliball,hong2024easyhec,zhang2026drhec} leverage robot motion and silhouette alignment, but applying them at scale remains challenging due to \emph{limited EEF visibility}, \emph{video--state temporal misalignment}, \emph{dual-arm base inconsistency}, and \emph{multi-camera settings}. We address these challenges via a unified three-stage calibration approach (Fig.~\ref{fig:cam_calib}). detailed below.

\paragraph{Hybrid initialization.}
We obtain the camera intrinsics $\mathbf{K}$ from official camera specifications or estimate them using MoGe-2~\cite{wang2025moge2} if unavailable. We initialize camera extrinsic parameters $(\mathbf{R},\mathbf{T})$ by minimizing a Perspective-n-Point (PnP) reprojection objective: $\min_{\mathbf{R},\mathbf{T}}
    \sum_{i}
    \left\|\pi_\mathbf{K}(\mathbf{R}\mathbf{p}_i+\mathbf{T})
    -\hat{\mathbf{p}}_i\right\|_2^2$
where $\mathbf p_i=(x_i,y_i,z_i)$ denotes a 3D point in robot space, $\hat{\mathbf p}_i=(u_i,v_i)$ its corresponding 2D pixel location, $\mathbf R$ and $\mathbf T$ denote the rotation and translation from robot space to camera space, respectively, and $\pi_\mathbf{K}$ denotes camera projection. To robustly solve it under challenging conditions, we employ the following two complementary strategies for 2D-3D correspondence acquisition.

\textit{Tracking the EEF origin point.}
This strategy is based on the observations of one physical point across multiple video frames. We first select a starting frame with high robot arm visibility in the image.
We then localize the EEF origin point by matching DINOv2~\cite{oquab2023dinov2} features against the pre-annotated point in a reference image of the gripper, following \cite{xie2025caliball}. Starting from this point, CoTracker3~\cite{karaev2024cotracker3} tracks the feature across video frames. We then compute the camera pose using these tracked correspondences and the robot-space EEF origin points obtained from robot states.

\textit{Tracking multiple points.}
To overcome the challenges of out-of-view EEF and limited gripper texture, we track multiple points on a visible rigid link with fixed relative geometry on the robot arm. At a starting video frame, we detect feature points within the arm mask and pose the URDF model into the camera space based on coarse camera estimates derived from installation configuration (we add random perturbations on camera pose and generate multiple candidates to account for the noise in the coarse estimate). For each feature whose viewing ray intersects a visible mesh triangle, we retain the intersection as a 3D model point $\mathbf{p}_i$ on the corresponding link. We then track these features to other frames using optical flow and solve camera pose via PnP.

We score the camera pose candidates computed from the above two strategies by the mean IoU between the projected URDF silhouette masks and arm/gripper masks detected by SAM3~\cite{carion2025sam3}, retaining the top-ranked candidate.

\begin{figure}[t!]
\centering
\includegraphics[width=\textwidth]{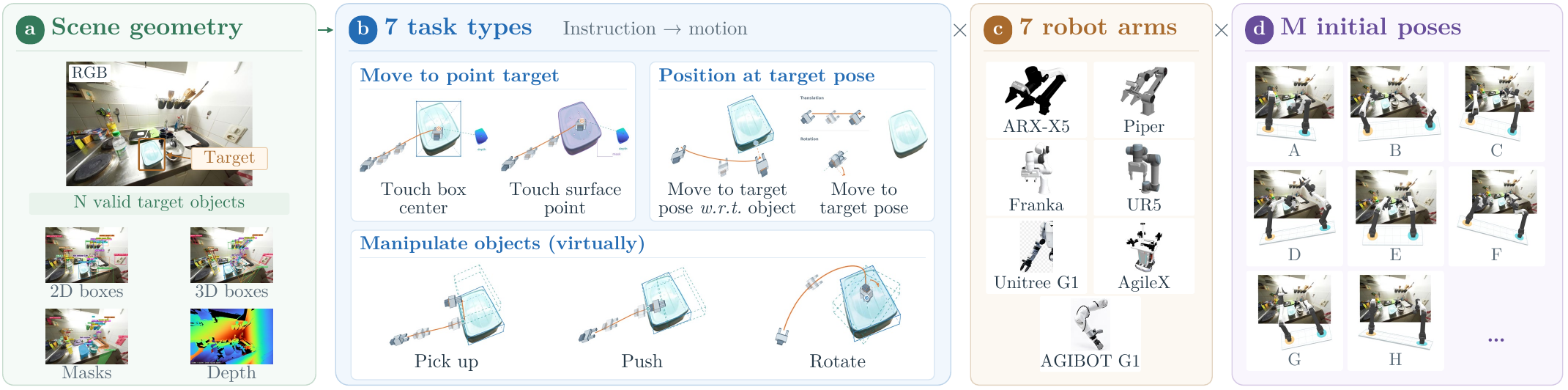}
\caption{\textbf{ImageAct: Synthesizing robot action targets from single images.}
Scene geometry localizes (a) $N$ objects. Combining these objects with (b) seven tasks, (c) seven robot embodiments, and (d) $M$ initial poses yields instruction-conditioned geometric action targets. Each combination yields a rendered image,
instruction, initial state, and action targets.}
\label{fig:img2act}
\end{figure}

\paragraph{Joint spatiotemporal refinement.}
We refine the initial calibration by aligning differentiable URDF renderings with observed robot masks. For a fixed, nonempty set $\mathcal{O}$ of camera, arm, and frame observations $(c,a,i)$, we optimize camera extrinsics $(\mathbf{R}_c,\mathbf{T}_c)$ alongside relative poses of arm bases $\mathbf{b}$ (omitted for single-arm setups) and joint time offsets $\boldsymbol\delta$ (to account for synchronization drift) by maximizing the soft intersection-over-union:
\begin{equation}
\label{eq:silhouette-refinement}
\begin{aligned}
    &\operatorname*{arg\,max}_{\{\mathbf{R}_c,\mathbf{T}_c\}, \mathbf{b}, \boldsymbol\delta}
    \frac{1}{|\mathcal{O}|}\sum_{(c,a,i)\in\mathcal{O}}
    \operatorname{IoU}_{\mathrm{s}}\!\left(M_{ci}^{a},\widehat M_{ci}^{a}\right),\\
    &\widehat M_{ci}^{a}=\mathcal{R}\!\left(\mathbf{K}_c,\mathbf{R}_c,\mathbf{T}_c,
    \operatorname{FK}\!\left(\mathcal{U},
    \mathbf{q}(\tau_{ci}^{a}+\delta_a),\mathbf{b}\right)\right).
\end{aligned}
\end{equation}
Here, $M_{ci}^{a}$ is the SAM3-detected arm mask, and $\mathcal{R}$ denotes the differentiable renderer. The forward kinematics operator $\operatorname{FK}$ articulates the unified URDF $\mathcal{U}$ after unifying robot base $\mathbf{b}$ for both arms and correcting the temporally shifted joint angles $\mathbf{q}(\tau_{ci}^{a}+\delta_a)$ computed at timestamp $\tau_{ci}^{a}$ and offset $\delta_a$.

To ensure robust convergence, we organize the optimization into a progressive three-step schedule. (1) Per-arm, per-camera refinement. We first refine the camera pose and temporal offset for each arm in each camera view by aligning rendered and observed arm masks. (2) Joint two-arm refinement. We then use the estimation from the previous step to recover the relative pose between the two arm bases, and transform their motion data into a unified robot reference coordinate system. Based on this unified motion, we jointly refine the shared camera pose and the relative base pose. (3) Multi-camera refinement. Building upon the single-view solutions, we jointly refine all camera poses and the shared relative base placement across multiple views, with temporal associations fixed. Refined candidates are validated by applying a binary mask IoU threshold against the SAM3-detected masks (analogous to candidate selection in initialization).

\paragraph{Calibration expansion.}
Many episodes fail validation across all candidates. To recover them, we leverage the fact that hardware setups stay relatively stationary across short sessions, so extrinsic parameters can be shared. We build a shared calibration pool by clustering reliable estimates into consensus candidates and then expand these candidates to unresolved episodes. Specifically, we either accept a consensus candidate directly if it passes IoU validation, or use the candidate to initialize the episode's calibration for further refinement.

\begin{table}[!t]
\centering
\caption{\textbf{ImageAct data source statistics.}}
\label{tab:imageact-data-sampling-combined}
\small
\setlength{\tabcolsep}{1.2pt}
\begin{tabular*}{\columnwidth}{@{\extracolsep{\fill}}lrrrr@{}}
\toprule
\multirow{2}{*}{Dataset} & \multicolumn{2}{c}{Images (M)} & \multicolumn{2}{c}{Objects (M)} \\
\cmidrule(lr){2-3}\cmidrule(lr){4-5}
& Before & After & Before & After \\
\midrule
CA1M (train) & 2.05 & 1.22 & 25.00 & 3.13 \\
Objects365 & 1.63 & 0.31 & 19.86 & 0.82\\
ScanNet++ & 3.92 & 2.80 & 511.52 & 9.61 \\
MesaTask & 23.55 & 19.56 & 159.55 & 64.42 \\
\midrule
\textbf{Total} & \textbf{31.15} & \textbf{23.89} & \textbf{715.93} & \textbf{77.98} \\
\bottomrule
\end{tabular*}
\end{table}

\subsection{ImageAct: Action synthesis from still images}
\label{sec:image2action}
\label{app:imageact-data}

Still images provide broad visual coverage but no executed robot motion.
ImageAct constructs action supervision from scene geometry by virtually placing robots within the scene and generating instruction-conditioned tasks (\cref{fig:img2act}). We use images from Objects365~\cite{shao2019objects365},
CA1M~\cite{lazarow2025ca1m}, ScanNet++~\cite{yeshwanth2023scannetpp}, and
MesaTask~\cite{mesatask2025}, spanning diverse natural objects, real indoor
environments, and synthetic tabletop scenes. After filtering, these sources
provide approximately \textbf{24M} images and \textbf{78M} object annotations. \Cref{tab:imageact-data-sampling-combined} presents the detailed image and object statistics.

\paragraph{Annotation.} For each image, we collect camera intrinsics and depth maps, together with
available object-level annotations: 2D and 3D bounding boxes, instance masks,
captions, and category labels. For Objects365, we augment the original 2D
boxes and category labels with offline predictions of metric depth and
camera intrinsics from MoGe-2~\cite{wang2025moge2}, instance masks from
SAM3~\cite{carion2025sam3}, and object captions from Qwen3-VL-8B-Instruct~\cite{bai2025qwen3}.
For CA1M, we use the provided 2D and 3D boxes, depth maps, and object captions.
For ScanNet++, we use the provided 3D boxes and derive per-image depth maps,
instance masks, and 2D boxes from annotated scene meshes and camera
parameters. MesaTask annotations come directly from synthetic scenes and
asset metadata.

\paragraph{Data filtering.}
We retain images with sufficient scene context and nearby objects, and
reject targets that are heavily occluded, substantially truncated by image
boundaries, or too small or too dominant in the image. We exclude objects
with sparse structures, such as flowers and wire baskets, for which 3D
bounding boxes alone cannot reliably specify interaction targets. We also
exclude categories unsuitable for robot manipulation, including walls,
floors, and people. Each target's category name or description must identify
it unambiguously within the image.

\paragraph{Tasks.}
For each sample, we render a virtual robot at its initial pose in the source
image and construct an instruction with a corresponding camera-space
trajectory for one of seven task types, organized into three groups
(\cref{fig:img2act}):
\begin{itemize}
\item \textbf{\emph{Move to point target}}: \emph{Touch box center} and \emph{Touch surface point} move the gripper
tip to a 3D point obtained by back-projecting a pixel $(u,v)$ using its depth
and the camera intrinsics. The former uses the target's 2D box center; the
latter uses a pixel specified in the instruction and selected within the
object's mask. The two tasks emphasize object identification and depth
estimation, respectively.
\item \textbf{\emph{Position at target pose}}: \emph{Move to target pose w.r.t. object} specifies the gripper's position and
orientation relative to the object's 3D box, using directions such as front,
left, right, and top. It emphasizes object pose understanding and diverse
end-effector rotations. \emph{Move to target pose} specifies translation or rotation by direction and
magnitude, allowing high-level agents to request fine end-effector
adjustments without referring to an object.
%\item \textbf{\emph{Manipulate objects (virtually)}}: \emph{Pick up}, \emph{Push}, and \emph{Rotate} combine approach and gripper closure with lifting, translation, and rotation, respectively, forming an extensible set of common manipulation primitives.
\item \textbf{\emph{Manipulate objects (virtually)}}:
\emph{Pick up}, \emph{Push}, and \emph{Rotate} move the gripper to a grasp pose derived from the object's 3D box, following the instructed approach direction. After closing, the gripper lifts against gravity,
pushes toward the object's center by a specified distance, or rotates
about the approach axis by a specified angle, respectively.
\end{itemize}
Converting geometric annotations into action targets makes this supervision directly compatible with the VLA's action prediction objective. The tasks
share spatial reasoning requirements and motion patterns with segments of
more complex robot tasks, providing a basis for transfer to downstream
manipulation.

\paragraph{Combinatorial expansion from one image.}
One image containing $N$ target objects can yield up to
$N\times7\times7\times M$ action-supervision samples by combining seven
task types, seven robot embodiments, and $M$ sampled initial arm positions.
For a fixed scene and initial robot configuration, instructions to approach
an object from the left or right specify different trajectories. The
instruction therefore determines the motion even when the visual input is
unchanged.

\paragraph{Box labels and supervision.}
The 2D and 3D bounding-box annotations collected for ImageAct can be applied to
supervise the box generation described in
\cref{sec:flow-matching}. For samples with these annotations, box prediction
provides auxiliary spatial supervision alongside the motion
targets.

\paragraph{Training sample generation.}
During training, ImageAct data samples are constructed online through the following steps:
\begin{itemize}
\item \emph{Image and object selection.} We sample an image from the four source datasets, then uniformly select a target from the objects retained after filtering.
\item \emph{Task and robot sampling.} We sample a task supported by the available object annotations and a robot embodiment for action generation and rendering.
\item \emph{Geometric augmentation.} We randomly sample the crop, output resolution, and aspect ratio, and add in-plane rotation when the sample does not require 2D bounding-box output. We keep the target within the image and update the camera intrinsics and object geometry accordingly~\cite{li2026scalable}.

\item \emph{Initial-state sampling.} We sample the initial arm-base poses, end-effector poses, gripper openings, and camera pose relative to the robot. Details of the sampling ranges and distributions are provided in Appendix~\ref{app:imageact}.

\item \emph{Action and instruction generation.} We sample task-specific parameters and use them to generate the action trajectory and construct the instruction from randomly sampled language templates. Details of action generation and instruction templates are provided in Appendix~\ref{app:imageact}.

\item \emph{Rendering and appearance augmentation.} We render the robot at its sampled initial configuration using the camera intrinsics of the geometrically augmented image and the sampled camera pose. After compositing the robot into the image, we apply random brightness, contrast, saturation, hue, and gamma adjustments, Gaussian noise, and JPEG compression.
\end{itemize}

\subsection{Egocentric Human Action Data}
Prior works \cite{li2026scalable,luo2026beingh05,yuan2026qwenrobotmanip,internw0} have demonstrated the effectiveness of incorporating egocentric human motion data to broaden scene and action coverage and support dexterous hand control. Following these studies, we incorporate such data as supplementary supervision for pretraining. Specifically, we use observation--instruction--hand-motion data processed from egocentric videos by \cite{li2026scalable} to train our model. In such data, human hands are treated as end effectors and represented using the MANO model~\cite{romero2017mano} with shape and joint-angle parameters. These samples use the corresponding valid dimensions in the shared interface.

\FloatBarrier

\section{Experimental Results}
\label{sec:experiments}
\label{app:experiments}
\subsection{Training and Evaluation Setup}
\label{sec:training}
\label{app:implementation}
\label{app:training-recipe}

\paragraph{Pretraining.}
For pretraining, we sample robot, ImageAct, and human data with a $15{:}3{:}1$ ratio. ImageAct samples are generated on the fly rather than drawn from a fixed set of precomputed trajectories (Sec.~\ref{sec:image2action}). We use the AdamW optimizer with $(\beta_1,\beta_2)=(0.9,0.95)$, weight decay $0.01$, and peak learning rates of $10^{-5}$ for the VLM and $10^{-4}$ for the action transformer. For the first 5,000 steps, the VLM learning rate is zero while the action transformer learning rate warms up linearly; both then reach their peak values and decay cosinusoidally to 10\% of them over the configured schedule. BF16 precision is used for training, and we sample 16 independent noise and flow-time values for each data example. We use a batch size of 2,048 and the exponential moving average (EMA) checkpoint at 150K steps to initialize the fine-tuning runs. Pretraining takes approximately 6 days on 32 NVIDIA B200 GPUs.

\paragraph{Fine-tuning.}
For adaptation, we finetune the model parameters on target-domain demonstrations, retaining the pretraining peak learning rates, BF16 precision, and 16 independent noise and flow-time samples per data example. The initial VLM freezing period and action warmup are shortened to 1,000 steps, followed by cosine decay to 10\% of the peak learning rates. EMA weights are used for evaluation. Fine-tuning time varies with dataset size and typically ranges from one to two days on 8 NVIDIA B200 GPUs for the evaluated benchmarks.

For simulation benchmark evaluation, we fine-tune our model on the benchmark training data and compare against prior methods using either their reported results or evaluations of their fine-tuned checkpoints. For real-robot experiments, we fine-tune all compared models on exactly the same data, including identical videos, language instructions, and language augmentations, and evaluate them after each has been trained on the same number of training samples.

\subsection{Inference Speed}
\label{app:inference}

At inference, we use 10 flow-matching steps to transform the initial noise into an action chunk, processing the concatenated sequence at each step.
Our model achieves a median inference latency of \textbf{93\,ms} per action chunk on a single NVIDIA GeForce RTX 5090 GPU using common optimization techniques including FlashAttention and CUDA graph optimization.

\begin{figure}[t!]
\centering
\includegraphics[width=\textwidth]{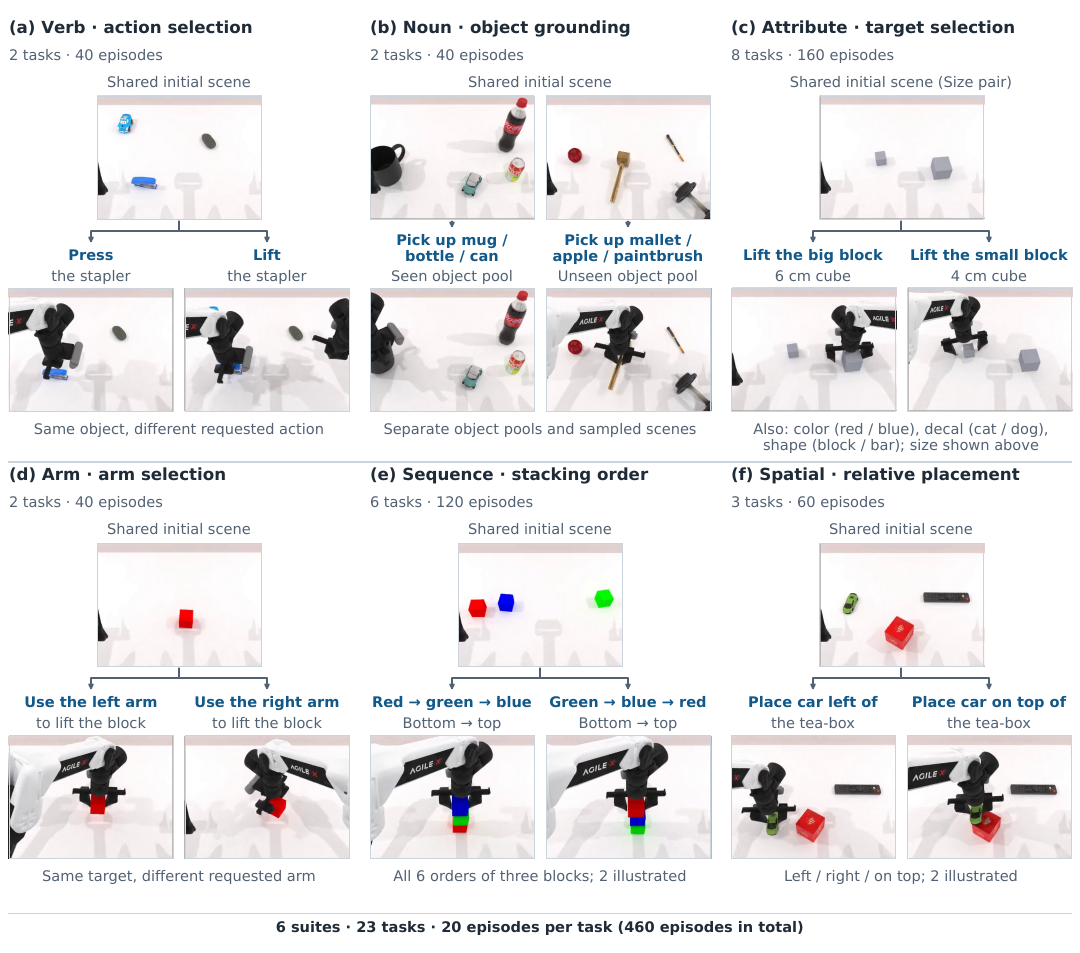}
\vspace{-20pt}
\caption{\textbf{RoboTwin-Instr benchmark task illustration.} The panels show initial scenes, example instructions, and successful outcomes for the six suites: (a) Verb, (b) Noun, (c) Attribute, (d) Arm, (e) Sequence, and (f) Spatial. See text for details of this benchmark.}
\label{fig:robotwin-if-tasks}
\end{figure}

\subsection{Simulation Benchmarks}
\label{sec:simulation}

\begin{table}[t!]
    \centering
    \small
    \caption{\textbf{Results on LIBERO benchmark series.} Average success rates (\%) are reported. Our model is evaluated five times on each benchmark. $^*$ denotes models without embodied pretraining; the best results are highlighted in bold. See Appendix~\ref{app:libero-results} for detailed comparisons on each benchmark.}
    \label{tab:libero-summary-chronological}
    \newcommand{\LiberoSummaryMethodWidth}{5cm}
    \newcommand{\LiberoSummarySizeWidth}{2cm}
    \newcommand{\LiberoSummaryResultWidth}{2.3cm}
    \setlength{\tabcolsep}{1pt}
    \renewcommand{\arraystretch}{1.0}
    \begin{adjustbox}{max width=\linewidth}
    \begin{threeparttable}[b]
    \begin{tabular}{@{}>{\raggedright\arraybackslash}p{\LiberoSummaryMethodWidth}>{\raggedright\arraybackslash}p{\LiberoSummarySizeWidth}*{3}{>{\raggedright\arraybackslash}p{\LiberoSummaryResultWidth}}@{}}
        \toprule
        Model & Size (B)\tnote{$\dagger$} & LIBERO & LIBERO-Plus~~~ & LIBERO-Pro \\
        \midrule
        $\pi_{0.5}$~\cite{black2025pi05} & 3.4 & 96.9 & 84.4 & 58.6 \\
        X-VLA~\cite{zheng2025xvla} & 0.9 & 98.1 & ~--- & 45.8 \\
        Spatial Forcing~\cite{li2026spatial} & 7.7 & ~--- & ~--- & 45.1 \\
        Motus~\cite{bi2026motus} & 8.2 & 97.7 & ~--- & ~--- \\
        Cosmos Policy$^*$~\cite{kim2026cosmospolicy} & 2.0 & 98.5 & 82.2 & 50.5 \\
        LingBot-VA~\cite{li2026lingbotva} & 5.5 & 98.5 & 73.6 & ~--- \\
        VLA-JEPA~\cite{sun2026vlajepa} & 2.3 & 97.2 & 79.5 & 51.0 \\
        ABot-M0~\cite{yang2026abotm0} & 5.6 & 98.6 & 80.5 & ~--- \\
        SimVLA$^*$~\cite{luo2026simvla} & 0.5 & 98.6 & ~--- & 48.6 \\
        Fast-WAM$^*$~\cite{yuan2026fastwam} & 6.2 & 97.6 & ~--- & 41.6 \\
        GR00T-N1.7~\cite{nvidia2026grootn17} & 3.1 & 97.0 & ~--- & ~--- \\
        MolmoAct2~\cite{fang2026molmoact2} & 5.0 & 97.2 & ~--- & ~--- \\
        Qwen-RobotManip~\cite{yuan2026qwenrobotmanip} & 4.9 & 99.1 & 89.0 & ~--- \\
        ImageWAM$^*$~\cite{zhang2026imagewam} & 4.5 & 98.4 & 83.1 & ~--- \\
        InternVLA-A1.5~\cite{internvla_a15} & 2.7 & ~--- & 84.8 & 55.4 \\
        4D-WAM$^*$~\cite{yang2026fourdwam} & 6.2 & 98.6 & 71.0 & 49.7 \\
        Flex-$\pi$~\cite{yan2026flexpi} & 6.3 & 98.5 & 88.6 & 50.6 \\
        VLAct~\cite{vlact2026} & 4.5 & ~--- & 82.6 & ~--- \\
        OpenWAM-$\alpha$~\cite{wang2026openwam} & 6.2 & 99.3 & 69.2 & 56.3 \\
        InternW0~\cite{cai2026internw0} & 7.6 & 98.6 & ~--- & ~--- \\
        InternW0-$\Delta$~\cite{internw0} & 8.6 & ~--- & 92.8 & ~--- \\
        UniWAM~\cite{song2026uniwam} & 8.2 & 99.2 & 92.6 & ~--- \\
        \midrule
        \textbf{Ours} & 2.4 & \textbf{99.4}{\tiny \,$\pm$0.15} & \textbf{94.6}{\tiny \,$\pm$0.24}  & \textbf{63.2}{\tiny \,$\pm$0.25} \\
        \bottomrule
    \end{tabular}
    \begin{tablenotes}
    \footnotesize
\item[$\dagger$] Throughout this paper, we report model size as the total number of parameters in all learned modules required for action inference, including frozen components but excluding standalone text encoders and training-only auxiliary modules.
\end{tablenotes}
    \end{threeparttable}
    \end{adjustbox}
\end{table}

\begin{table}[htbp]
\vspace{-10pt}
    \centering
    \footnotesize
    \setlength{\tabcolsep}{0.8pt}
    \caption{\textbf{RoboTwin 2.0 benchmark results.}
        Success rates (\%) under clean-to-random (clean-only training) and clean+random settings are reported. $^*$ denotes models without embodied pretraining.}
    \label{tab:robotwin-combined}
    \newcommand{\RoboTwinMethodWidth}{4.1cm}
    \newcommand{\RoboTwinSizeWidth}{1.2cm}
    \newcommand{\RoboTwinResultWidth}{1.25cm}
    \vspace{-5pt}
    \begin{adjustbox}{max width=\linewidth}
    \begin{tabular}{@{}>{\raggedright\arraybackslash}p{\RoboTwinMethodWidth}>{\raggedright\arraybackslash}p{\RoboTwinSizeWidth}>{\centering\arraybackslash}p{\RoboTwinResultWidth}>{\columncolor{reportaccent!8}\centering\arraybackslash}p{\RoboTwinResultWidth}*{3}{>{\centering\arraybackslash}p{\RoboTwinResultWidth}}>{\columncolor{reportaccent!8}\centering\arraybackslash}p{\RoboTwinResultWidth}@{}}
        \toprule
        & & \multicolumn{3}{c}{Clean-only train.} & \multicolumn{3}{c}{Clean+random} \\
        \cmidrule(lr){3-5}\cmidrule(l){6-8}
        Model & (B) & Cln. & \textbf{Rnd.} & Avg. & Cln. & Rnd. & \textbf{Avg.} \\
        \midrule
        $\pi_{0.5}$~\cite{black2025pi05} & 3.4 & 73.1 & 47.9 & 60.5 & 82.7 & 76.8 & 79.8 \\
        X-VLA~\cite{zheng2025xvla} & 0.9 & 68.0 & 20.9 & 44.5 & --- & --- & --- \\
        Spatial Forcing~\cite{li2026spatial} & 7.7 & 77.2 & 26.7 & 52.0 & --- & --- & --- \\
        Motus~\cite{bi2026motus} & 8.2 & --- & --- & --- & 88.7 & 87.0 & 87.8 \\
        LingBot-VLA~\cite{wu2026lingbot} & 4.2 & --- & --- & --- & 88.6 & 86.7 & 87.6 \\
        LingBot-VA~\cite{li2026lingbotva} & 5.5 & --- & --- & --- & 92.9 & 91.5 & 92.2 \\
        ABot-M0~\cite{yang2026abotm0} & 5.6 & 70.7 & 36.0 & 53.4 & 86.1 & 85.1 & 85.6 \\
        Fast-WAM$^*$~\cite{yuan2026fastwam} & 6.2 & 77.8 & ~\,1.9 & 39.9 & 91.9 & 91.8 & 91.8 \\
        GR00T-N1.7~\cite{nvidia2026grootn17} & 3.1 & 43.6 & 20.7 & 32.2 & --- & --- & --- \\
        X-WAM~\cite{guo2026xwam} & 5.2 & 70.0 & 25.8 & 47.9 & --- & --- & --- \\
        Being-H0.7~\cite{luo2026beingh07} & --- & --- & --- & --- & 90.2 & 89.6 & 89.9 \\
        AHA-WAM$^*$~\cite{cai2026ahawam} & 7.5 & 64.3 & ~\,3.2 & 33.8 & 93.4 & 92.2 & 92.8 \\
        Hy-Embodied-0.5~\cite{zhang2026hyvla} & 4.5 & --- & --- & --- & 90.9 & 90.1 & 90.5 \\
        Qwen-RobotManip~\cite{yuan2026qwenrobotmanip} & 4.9 & 84.7 & 69.4 & 77.1 & 93.7 & 94.0 & 93.9 \\
        ABot-M0.5~\cite{chen2026abotm05} & 5.0 & --- & --- & --- & 94.0 & 94.2 & 94.1 \\
        LingBot-VA 2.0~\cite{zhang2026lingbotva2} & \!\!\!13.6 (2.5) & --- & --- & --- & 93.8 & 93.4 & 93.6 \\
        4D-WAM$^*$~\cite{yang2026fourdwam} & 6.2 & 81.5 & 41.8 & 61.7 & --- & --- & --- \\
        GigaBrain-0.7-Base~\cite{gigabrain2026v07}\!\!\!\! & 3.5 & 66.8 & 67.9 & 67.4 & --- & --- & --- \\
        VLAct~\cite{vlact2026} & 4.5 & 80.5 & --- & --- & 92.5 & 90.8 & 91.7 \\
        OpenWAM-$\alpha$~\cite{wang2026openwam} & 6.2 & 89.4 & 48.7 & 69.0 & 93.7 & 93.5 & 93.6 \\
        InternW0~\cite{cai2026internw0} & 7.6 & --- & --- & --- & 93.2 & 93.0 & 93.1 \\
        InternW0-$\Delta$~\cite{internw0} & 8.6 & 90.0 & 71.9 & 81.0 & --- & --- & --- \\
        UniWAM~\cite{song2026uniwam} & 8.2 & 75.1 & 68.3 & 71.7 & --- & --- & ---\\
        \midrule
        \textbf{Ours} & 2.4 & \textbf{90.4} & \textbf{74.0} & \textbf{82.2} & \textbf{95.9} & \textbf{94.8} & \textbf{95.3} \\
        \bottomrule
    \end{tabular}
    \end{adjustbox}

\end{table}
\begin{table}[htbp]
    \centering
    \footnotesize
    \caption{\textbf{RoboDojo benchmark results.} Model size (B), success rates (\%), and progress-based scores.}
    \label{tab:robodojo-policy}
    \newcommand{\RoboDojoMethodWidth}{4.3cm}
    \newcommand{\RoboDojoSizeWidth}{0.7cm}
    \newcommand{\RoboDojoResultWidth}{1.1cm}
    \setlength{\tabcolsep}{1.5pt}
    \vspace{-5pt}
    \begin{adjustbox}{max width=\linewidth}
    \begin{tabular}{@{}>{\raggedright\arraybackslash}p{\RoboDojoMethodWidth}>{\raggedright\arraybackslash}p{\RoboDojoSizeWidth}>{\raggedright\arraybackslash}p{\RoboDojoResultWidth}>{\raggedright\arraybackslash}p{\RoboDojoResultWidth}@{}}
        \toprule
        Model & (B) & ~~~~SR\!\! & {Score} \\
        \midrule
        $\pi_{0.5}$~\cite{black2025pi05} & 3.4 & ~\,6.93 & 11.44 \\
        LingBot-VLA~\cite{wu2026lingbot} & 4.2 & ~\,2.96 & ~\,5.50 \\
        Xiaomi-Robotics-0~\cite{cai2026xiaomi} & 4.7 & ~\,4.18 & ~\,6.93 \\
        StarVLA-OFT~\cite{starvla2025} & 4.5 & ~\,4.82 & ~\,8.18 \\
        X-WAM~\cite{guo2026xwam} & 5.2 & ~\,3.83 & ~\,7.69 \\
        GalaxeaVLA G0.5~\cite{liu2026g05} & 2.3 & 14.88 & 20.23 \\
        Hy-Embodied-0.5~\cite{zhang2026hyvla} & 4.5 & ~\,8.80 & 13.07 \\
        InternVLA-A1.5~\cite{internvla_a15} & 2.7 & ~\,7.14 & 11.15 \\
        DM0.5~\cite{dexmal2026dm05} & 5.2 & 19.34 & 24.90 \\
        VLAct~\cite{vlact2026} & 4.5 & ~\,7.58 & 10.65 \\
        OpenWAM-$\alpha$~\cite{wang2026openwam} & 6.2 & 11.92 & 17.18 \\
        InternW0-$\Delta$~\cite{internw0} & 8.6 & 23.91 & 30.77 \\
        VPP2~\cite{guo2026vpp2} & \!\!\!19.1 & \textbf{32.26} & \textbf{39.26} \\
        \midrule
        \textbf{Ours} & 2.4 & 26.53 & 32.24 \\
        \bottomrule
    \end{tabular}
    \end{adjustbox}
\end{table}

\begin{table}[htbp]
    \centering
    \footnotesize
    \caption{\textbf{Robotwin-Instr benchmark results.} We introduce RoboTwin-Instr to further probe generalization and instruction understanding (see text for details). Model size (B) and success rates (\%) are reported. Models trained with  the  clean+random mixed training setting are used here for evaluation.}
    \label{tab:robotwin-if-main}
    \newcommand{\RoboTwinIFMethodWidth}{4.1cm}
    \newcommand{\RoboTwinIFSizeWidth}{0.75cm}
    \newcommand{\RoboTwinIFResultWidth}{1.2cm}
    \setlength{\tabcolsep}{0.8pt}
    \vspace{-5pt}
    \begin{adjustbox}{max width=\linewidth}
    \begin{tabular}{@{}>{\raggedright\arraybackslash}p{\RoboTwinIFMethodWidth}>{\raggedright\arraybackslash}p{\RoboTwinIFSizeWidth}*{6}{>{\centering\arraybackslash}p{\RoboTwinIFResultWidth}}>{\columncolor{reportaccent!8}\centering\arraybackslash}p{\RoboTwinIFResultWidth}@{}}
        \toprule
        Model & (B) & Verb & Noun & Attr. & Arm & Seq. & Spat. & \textbf{Avg.} \\
        \midrule
        Motus~\cite{bi2026motus} & 8.2 & 92.5 & 65.0 & 52.5 & 65.0 & 6.7 & 36.7 & 53.1 \\
        LingBot-VLA~\cite{wu2026lingbot} & 4.2 & 47.5 & 32.5 & 48.1 & 45.0 & 11.7 & 5.0 & 31.6 \\
        LingBot-VA~\cite{li2026lingbotva} & 5.5 & 45.0 & 75.0 & 67.5 & \textbf{92.5} & \textbf{14.2} & \textbf{43.3} & 56.3 \\
        Fast-WAM~\cite{yuan2026fastwam} & 6.2 & 95.0 & 65.0 & 56.3 & 47.5 & 8.3 & 38.3 & 51.7 \\
        X-WAM~\cite{guo2026xwam} & 5.2 & 85.0 & 72.5 & 50.6 & 10.0 & 5.8 & 30.0 & 42.3 \\
        Hy-Embodied-0.5~\cite{zhang2026hyvla}\!\!\! & 4.5 & 87.5 & 27.5 & 51.9 & 42.5 & 12.5 & 11.7 & 38.9 \\
        DM0.5~\cite{dexmal2026dm05} & 5.2 & 80.0 & 62.5 & 55.0 & 62.5 & 7.5 & 33.3 & 50.1 \\
        VLAct~\cite{vlact2026} & 4.5 & 90.0 & 70.0 & \textbf{68.1} & 72.5 & 12.5 & 31.7 & 57.5 \\
        \midrule
        \textbf{Ours} & 2.4 & \textbf{100.0} & \textbf{82.5} & 63.1 & 72.5 & 10.8 & 38.3 & \textbf{61.2} \\
        \bottomrule
    \end{tabular}
    \end{adjustbox}

\end{table}

\paragraph{Existing benchmarks.}
We evaluate our method on LIBERO~\cite{liu2023libero}, LIBERO-Plus~\cite{liberoplus2025}, LIBERO-Pro~\cite{zhou2025liberopro}, RoboTwin~2.0~\cite{chen2025robotwin2}, and RoboDojo~\cite{chen2026robodojo}, following the official training and evaluation protocols. For LIBERO-Pro, we follow existing methods to evaluate the first four perturbation types, excluding environment perturbations due to their instability. For RoboTwin 2.0, we evaluate two training settings: 50 clean demonstrations per task (Clean-only), and 50 clean $+$ 500 randomized demonstrations per task (Clean+Random). More details about these benchmarks and evaluations are provided in Appendix~\ref{app:simulation-results}.

\paragraph{A new benchmark: RoboTwin-Instr.} To further evaluate generalization and instruction understanding, we introduce \textbf{RoboTwin-Instr}, an evaluation benchmark built on RoboTwin 2.0. The benchmark design draws partial inspiration from \cite{yuan2026qwenrobotmanip}, while extending generalization evaluation across multiple dimensions, including object attributes, unseen objects, arm selection, and operation sequences. It covers six suites, including \emph{Verb} (action selection), \emph{Noun} (object-name grounding), \emph{Attribute} (attribute-based selection), \emph{Arm (arm selection)}, \emph{Sequence} (stacking order), and \emph{Spatial} (relative placement). Figure~\ref{fig:robotwin-if-tasks} illustrates these suites, with further details provided in Appendix~\ref{app:robotwin-if}. A central design principle is to \emph{vary the instruction while keeping non-linguistic initial conditions fixed}. The benchmark contains 23 tasks and 460 testing episodes in total, and we evaluate models trained under the original RoboTwin 2.0 settings.

\paragraph{Results.}
Our model achieves strong performance across simulation benchmarks, outperforming many substantially larger models. It achieves 99.4\%, 94.6\%, and 63.2\% success on LIBERO, LIBERO-Plus, and LIBERO-Pro, respectively, leading all three comparisons (\cref{tab:libero-summary-chronological}). On RoboTwin~2.0, it achieves 74.0\% under clean-to-random transfer and a 95.3\% average across clean and randomized evaluations with mixed training, also leading both comparisons (\cref{tab:robotwin-combined}). On RoboDojo, it outperforms other models such as the 8.6B InternW0-$\Delta$ while trailing the recent 19.1B VPP2 model (\cref{tab:robodojo-policy}). On RoboTwin-Instr evaluated with the RoboTwin Clean+Random mixed training models, ours achieves the highest average success rate of 61.2\%, outperforming the next-best model VLAct by 3.7 percentage points (\cref{tab:robotwin-if-main}). It leads on action selection and object-name grounding, while remaining competitive on stacking order and relative placement. These results highlight strong manipulation capabilities and robustness to distribution shifts within a compact pretrained architecture.

\begin{table}[t!]
  \centering
  \caption{\textbf{Fine-tuning data statistics for the AGIBOT G1 tasks.}}
  \label{tab:training_data}
  \small
  \begin{adjustbox}{max width=\linewidth}
  \begin{tabular}{lrrr}
    \toprule
    Skills              & Episodes &    Frames & Hours \\
    \midrule
    Pick/Pass/Place Objects & 14{,}717 & 2{,}343{,}242 & 21.7 \\
    Pour Water (w. kettles)        &  2{,}910 & 1{,}015{,}997 &  9.4 \\
    Pour Tea (w. teapots)        &  4{,}137 & 1{,}454{,}273 & 13.5 \\
    Other skills      &  8{,}861 & 2{,}110{,}356 & 19.5 \\
    \midrule
    Total             & 30{,}625 & 6{,}923{,}868 & 64.1 \\
    \bottomrule
  \end{tabular}
  \end{adjustbox}
\end{table}

\begin{table}[t!]
  \centering
  \caption{\textbf{Fine-tuning data statistics for AGIBOT G2 and WUJI-1 dexterous-hand tasks.}}
  \label{tab:adaptation_data}
  \small
  \setlength{\tabcolsep}{6pt}
  \begin{adjustbox}{max width=\linewidth}
  \begin{tabular}{llrr}
    \toprule
    Platform & Task & Episodes & Hours \\
    \midrule
    AGIBOT G2 + OmniPicker & Laundry unloading & 898 & 7.3 \\
    AGIBOT G2 + DH gripper (custom fingertips) & Workpiece loading & 2{,}299 & 18.2 \\
    \midrule
    WUJI-1 dexterous hands + Tianji Marvin arms & Bottle-cap unscrewing & 726 & 3.0 \\
    WUJI-1 dexterous hands + Tianji Marvin arms & Playing-card pickup   & 936 & 1.1 \\
    \bottomrule
  \end{tabular}
  \end{adjustbox}
\end{table}

\subsection{Real-Robot Experiments}
\label{sec:real-setup}
\label{app:real-robot-setup}
\label{app:real-generalization-setup}
\label{app:real-hardware}

We evaluate our model on diverse real-world manipulation tasks across three robot platforms: AGIBOT G1 (our primary testbed), AGIBOT G2 (with two gripper configurations), and WUJI-1 dexterous hands mounted on Tianji Marvin arms (Fig.~\ref{fig:real-rollouts}).

\subsubsection{Experimental Setup}

\begin{figure}[htbp]
\centering
\captionsetup{
    justification=raggedright,
    singlelinecheck=false
}
\captionsetup[subfigure]{
    justification=raggedright,
    singlelinecheck=false,
    labelfont=bf
}

\noindent
\begin{subfigure}{\textwidth}
    \includegraphics[width=\linewidth]
        {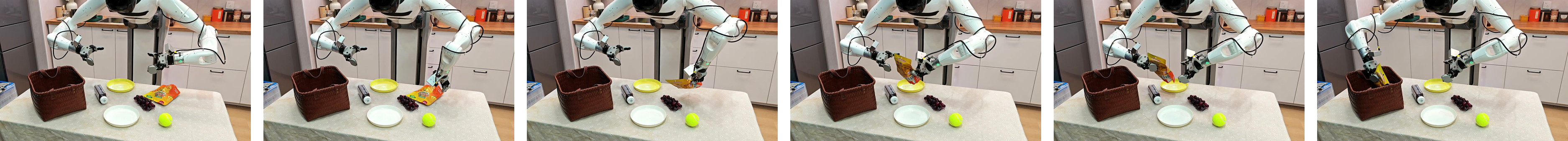}
    \caption{\textbf{Pick--Pass--Place (AGIBOT G1).}
        \textit{``Pick up the orange-yellow candy bag with the left arm.''}
        $\rightarrow$
        \textit{``Pass the orange-yellow candy bag to the right arm
        with the left arm.''}
        $\rightarrow$
        \textit{``Place the orange-yellow candy bag into the brown basket
        with the right arm.''}}
    \label{fig:pick_pass_place}
\end{subfigure}

\par\smallskip
\noindent
\begin{subfigure}{\textwidth}
    \includegraphics[width=\linewidth]
        {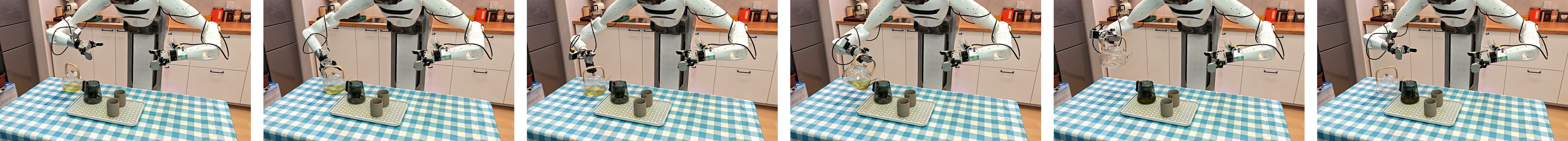}
    \caption{\textbf{Pour water (AGIBOT G1).}
        \textit{``Pick up the kettle from the table by its handle
        with the right arm.''}
        $\rightarrow$
        \textit{``Pour water from the kettle into the teapot on
        the tea tray with the right arm.''}
        $\rightarrow$
        \textit{``Place the kettle at the front right of the table
        with the right arm.''}}
    \label{fig:pour_water}
\end{subfigure}

\par\smallskip
\noindent
\begin{subfigure}{\textwidth}
    \includegraphics[width=\linewidth]
        {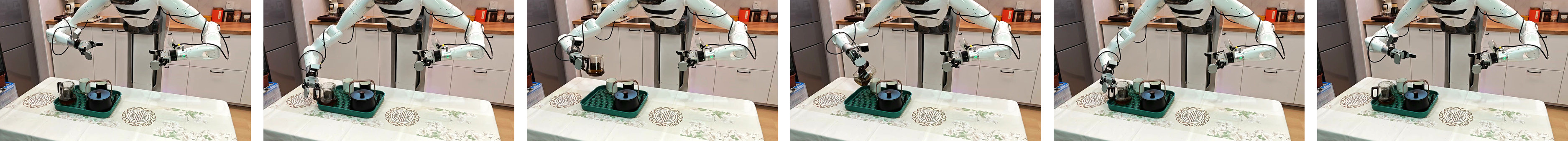}
    \caption{\textbf{Pour tea (AGIBOT G1).}
        \textit{``Pick up the teapot from the tea tray by its handle
        with the right arm.''}
        $\rightarrow$
        \textit{``Pour tea from the teapot into the right cup on
        the tea tray with the right arm.''}
        $\rightarrow$
        \textit{``Place the teapot at the back right of the tea tray
        with the right arm.''}}
    \label{fig:pour_tea}
\end{subfigure}

\par\smallskip
\noindent
\begin{subfigure}{\textwidth}
    \includegraphics[width=\linewidth]
        {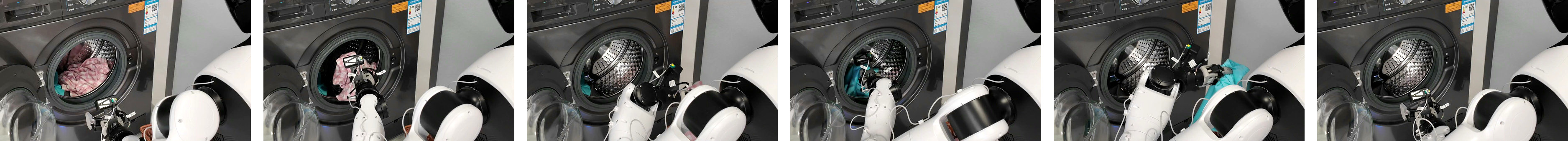}
    \caption{\textbf{Laundry unloading (AGIBOT G2).}
        \textit{``Use the left arm to reach into the washing machine
        and grasp a piece of clothing, then bring it out and place
        it in the laundry basket held by the right arm.''}}
    \label{fig:laundry}
\end{subfigure}

\par\smallskip
\noindent
\begin{subfigure}{\textwidth}
    \includegraphics[width=\linewidth]
        {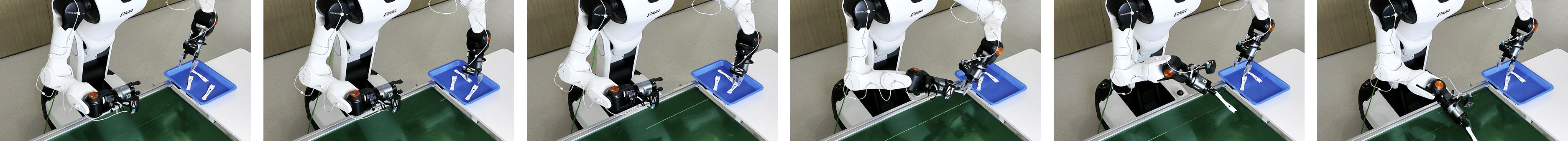}
    \caption{\textbf{Workpiece loading (AGIBOT G2 w. a custom gripper).}
        \textit{``Use the left arm to pick up a medical item from
        the material tray on the left and hand it over to the right
        arm, then use the right arm to receive the item and place
        it on the conveyor belt.''}}
    \label{fig:medical_loading}
\end{subfigure}

\par\smallskip
\noindent
\begin{subfigure}{\textwidth}
    \includegraphics[width=\linewidth]
        {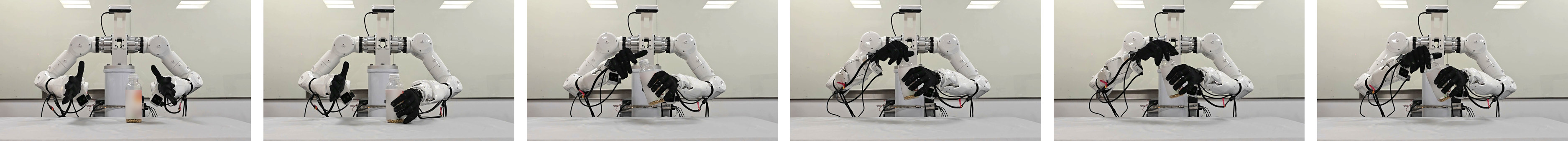}
    \caption{\textbf{Bottle-cap unscrewing (WUJI-1 dexterous hands on Tianji arms).}
        \textit{``Left hand: Pick up and stabilize the bottle.
        Right hand: Hold position, then approach the bottle as the
        left hand brings it into position.''}
        $\rightarrow$
        \textit{``Left hand: Hold the bottle.
        Right hand: Fully unscrew the cap.''}}
    \label{fig:unscrew_bottle_cap}
\end{subfigure}

\par\smallskip
\noindent
\begin{subfigure}{\textwidth}
    \includegraphics[width=\linewidth]
        {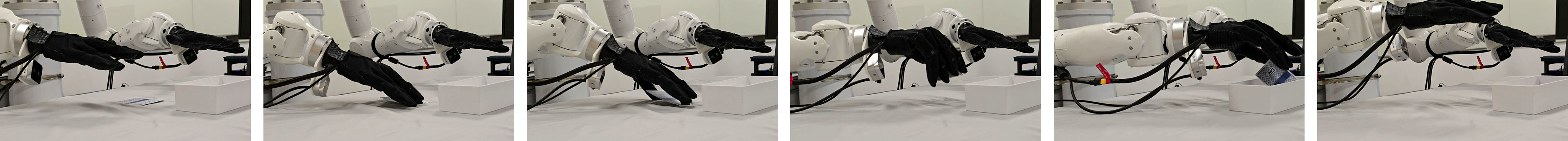}
    \caption{\textbf{Playing-card pickup (WUJI-1 dexterous hands on Tianji arms).}
        \textit{``Left hand: Hold position. Right hand: Press down on the edge
        of the playing card with the right index finger to make the edge tilt up.''}
        $\rightarrow$
        \textit{``... Pinch the upturned edge with the right thumb.''}
        $\rightarrow$
        \textit{``... Put the playing card into the white box.''}}
    \label{fig:dex-card}
\end{subfigure}

\caption{\textbf{Real-robot tasks.}
    Examples of task execution across the tasks on AGIBOT G1, AGIBOT G2,
    and WUJI-1 dexterous hands with Tianji arms. Note that the two pouring tasks require different grasp poses and rotational motions during pouring because the vessels have different handle configurations.}
\label{fig:real-rollouts}
\end{figure}

\paragraph{AGIBOT G1 tasks.}
We use an AGIBOT G1 robot equipped with an OmniPicker gripper as our primary testbed to evaluate multi-task accuracy and generalization. 
In \emph{Pick--Place}, the robot grasps a designated object with one arm, and then places it at a specified spatial relation relative to a reference object. In \emph{Pick--Pass--Place}, the robot needs to additionally pass the grasped object to the other arm before placing it (Fig.~\ref{fig:real-rollouts}a). In \emph{Pour Water} and \emph{Pour Tea}, the robot grasps a vessel (a kettle with top handles and a teapot with a side handle, respectively) by its handle, aligns its spout to pour into a target container, and returns the vessel to a designated location (Fig.~\ref{fig:real-rollouts}b--c). Note that the two pouring tasks require different grasp poses and rotational motions during pouring due to their different handle configurations.

\begin{figure}[t!]
  \centering
  \begin{subfigure}[t]{0.16\textwidth}
    \includegraphics[width=\linewidth]{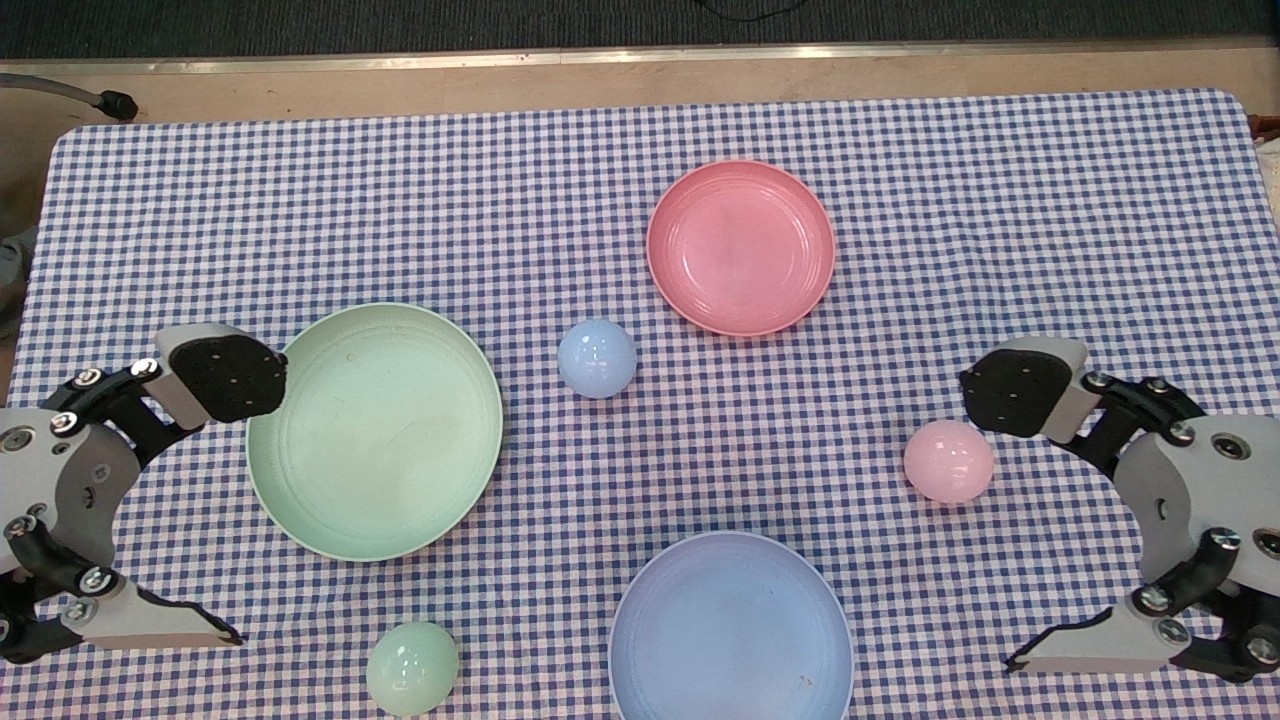}
    \caption{Balls 1}
  \end{subfigure}\hfill
  \begin{subfigure}[t]{0.16\textwidth}
    \includegraphics[width=\linewidth]{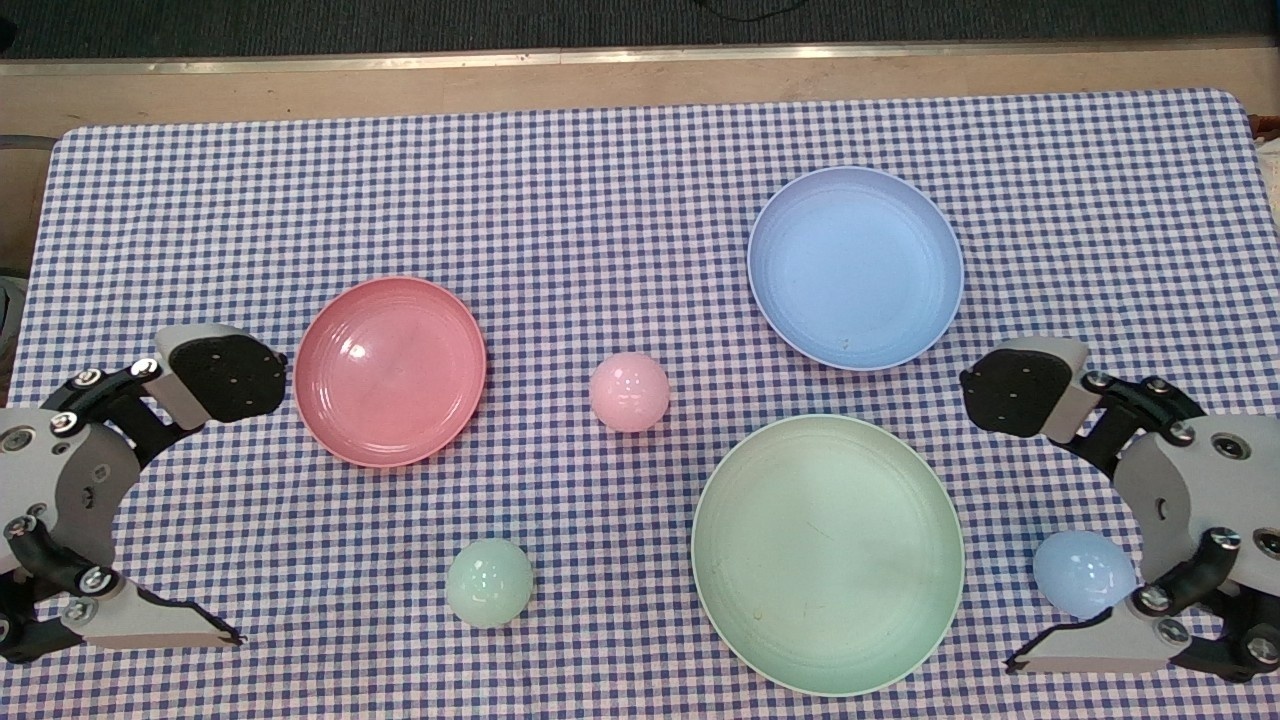}
    \caption{Balls 2}
  \end{subfigure}\hfill
  \begin{subfigure}[t]{0.16\textwidth}
    \includegraphics[width=\linewidth]{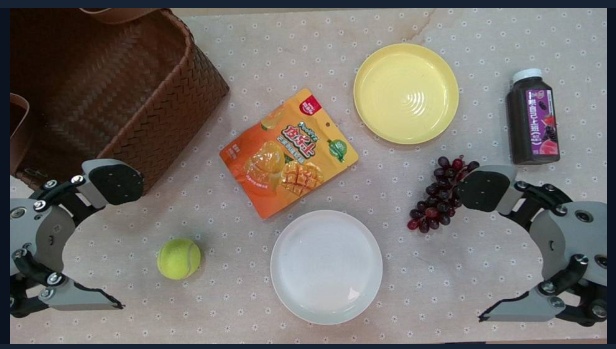}
    \caption{Clutter 1}
  \end{subfigure}\hfill
  \begin{subfigure}[t]{0.16\textwidth}
    \includegraphics[width=\linewidth]{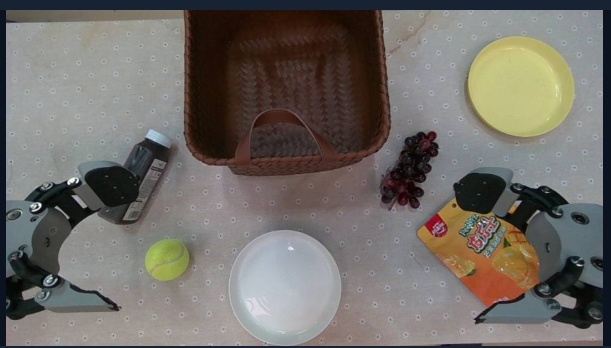}
    \caption{Clutter 2}
  \end{subfigure}\hfill
  \begin{subfigure}[t]{0.16\textwidth}
    \includegraphics[width=\linewidth]{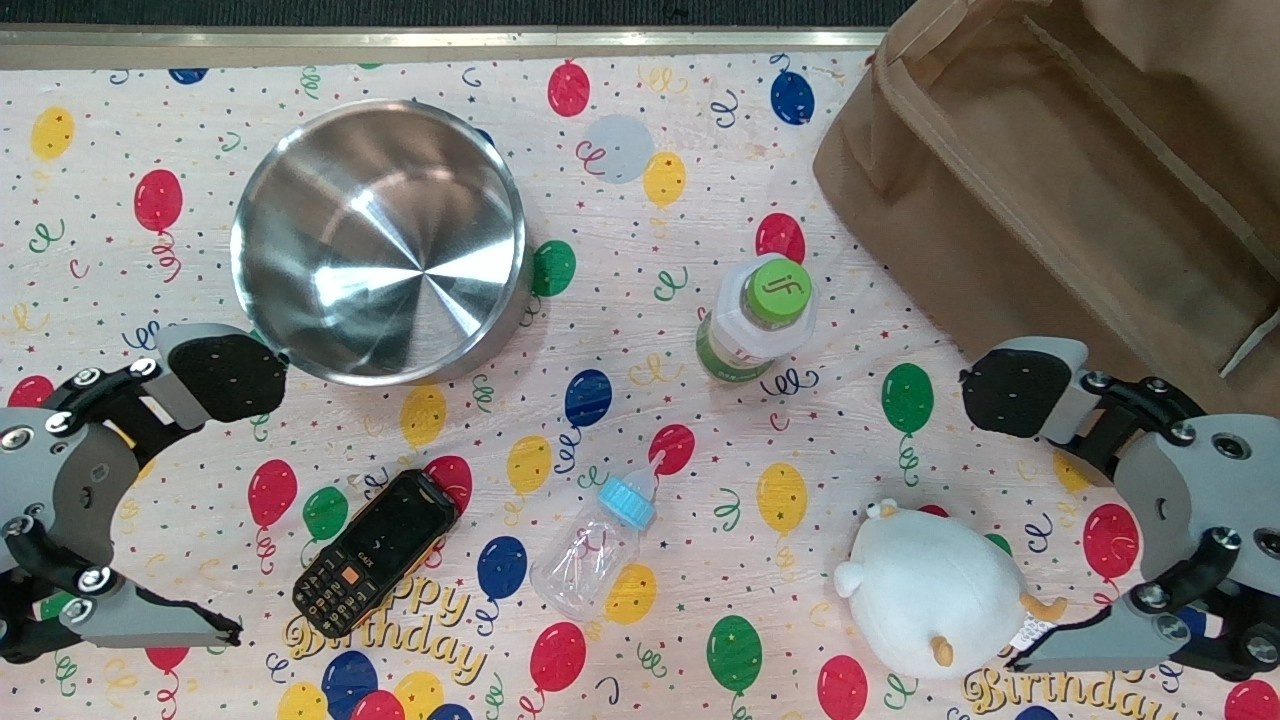}
    \caption{Unseen 1}
  \end{subfigure}\hfill
  \begin{subfigure}[t]{0.16\textwidth}
    \includegraphics[width=\linewidth]{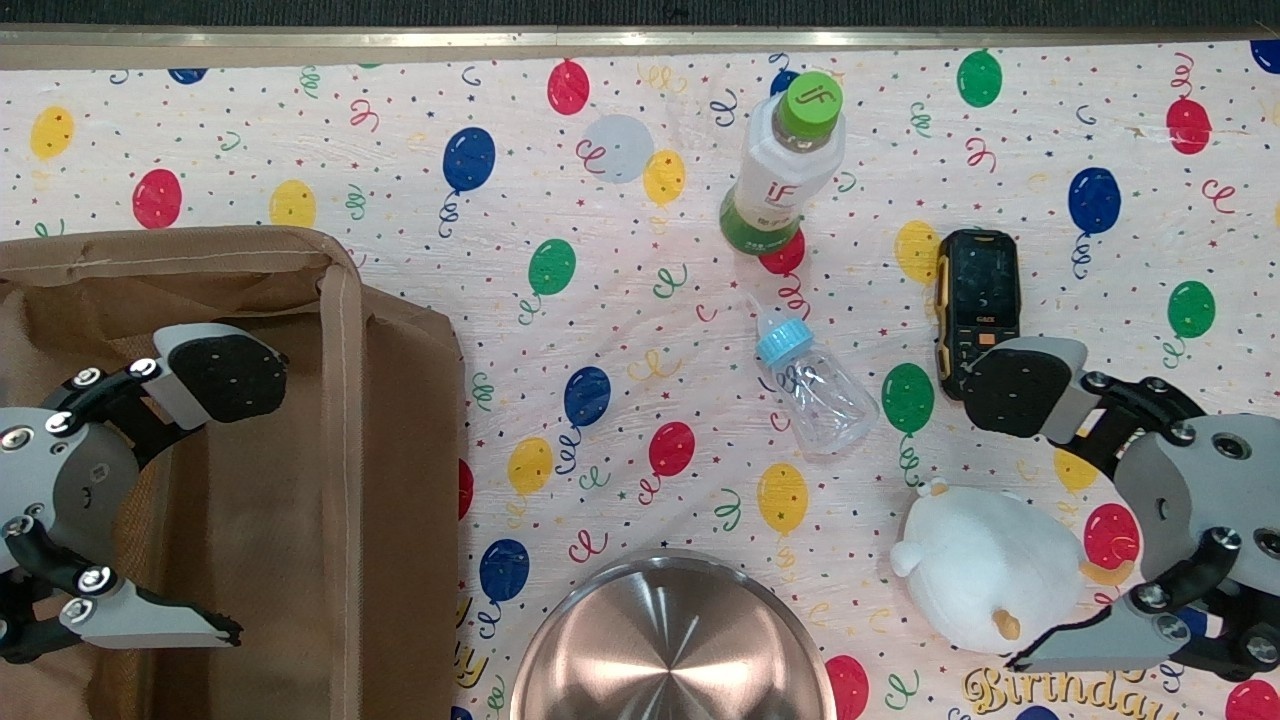}
    \caption{Unseen 2}
  \end{subfigure}

  \vspace{0.8ex}

  \begin{subfigure}[t]{0.16\textwidth}
    \includegraphics[width=\linewidth]{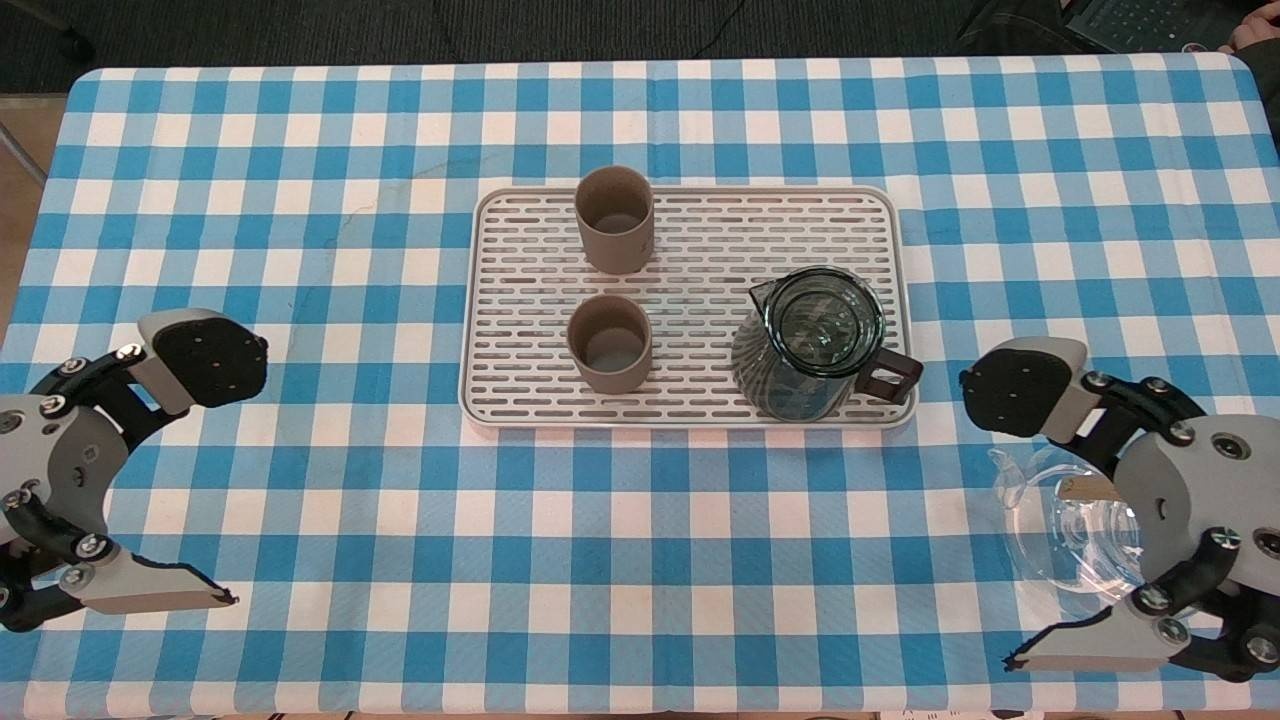}
    \caption{Pouring 1}
  \end{subfigure}\hfill
  \begin{subfigure}[t]{0.16\textwidth}
    \includegraphics[width=\linewidth]{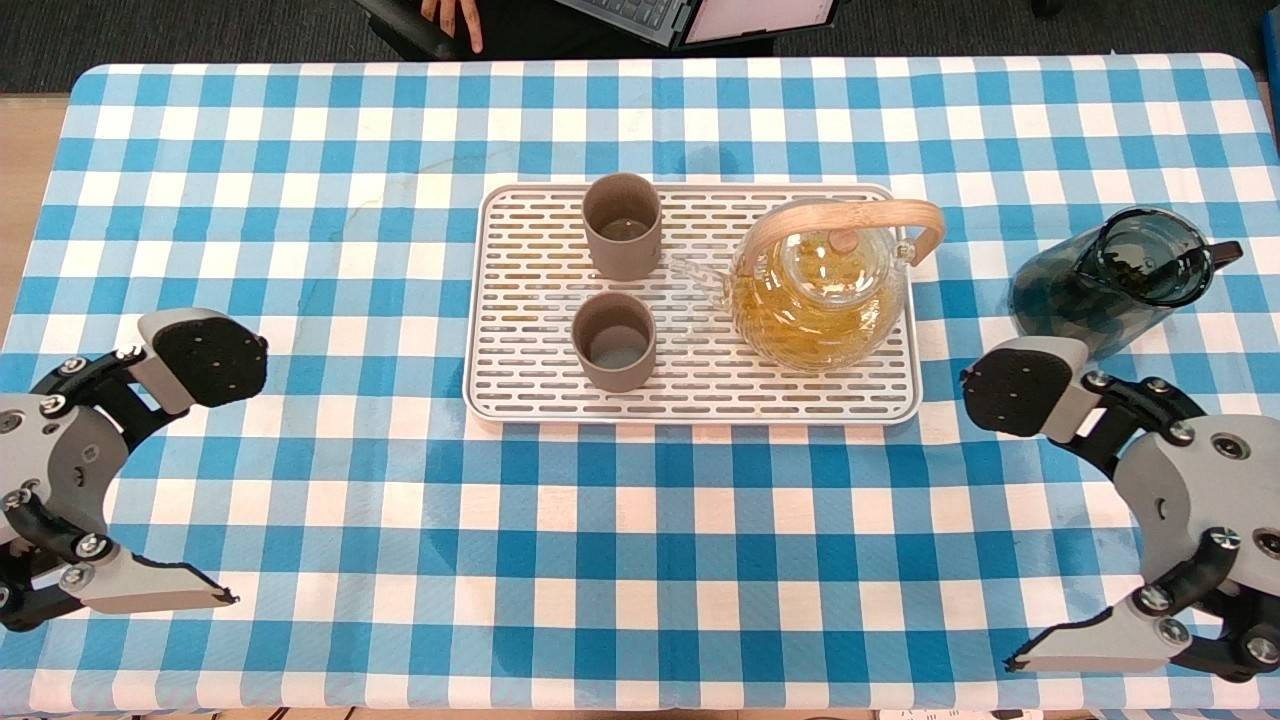}
    \caption{Pouring 2}
  \end{subfigure}\hfill
  \begin{subfigure}[t]{0.16\textwidth}
    \includegraphics[width=\linewidth]{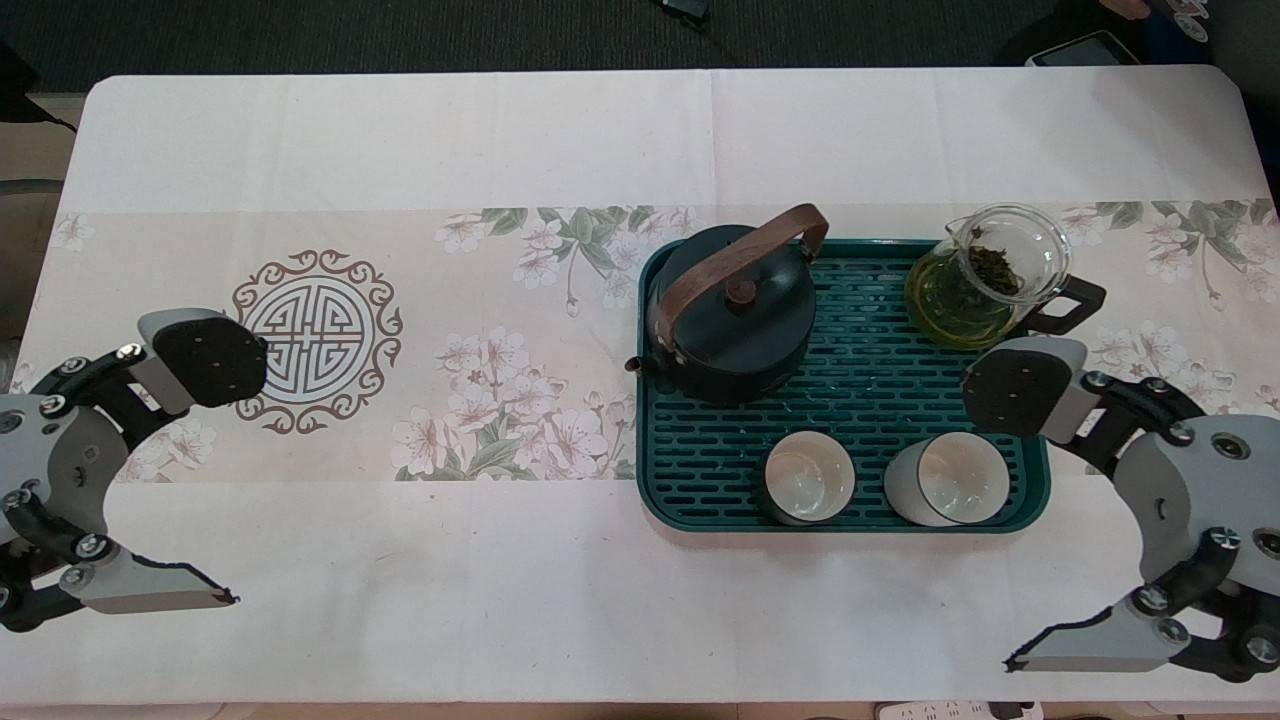}
    \caption{Pouring 3}
  \end{subfigure}\hfill
  \begin{subfigure}[t]{0.16\textwidth}
    \includegraphics[width=\linewidth]{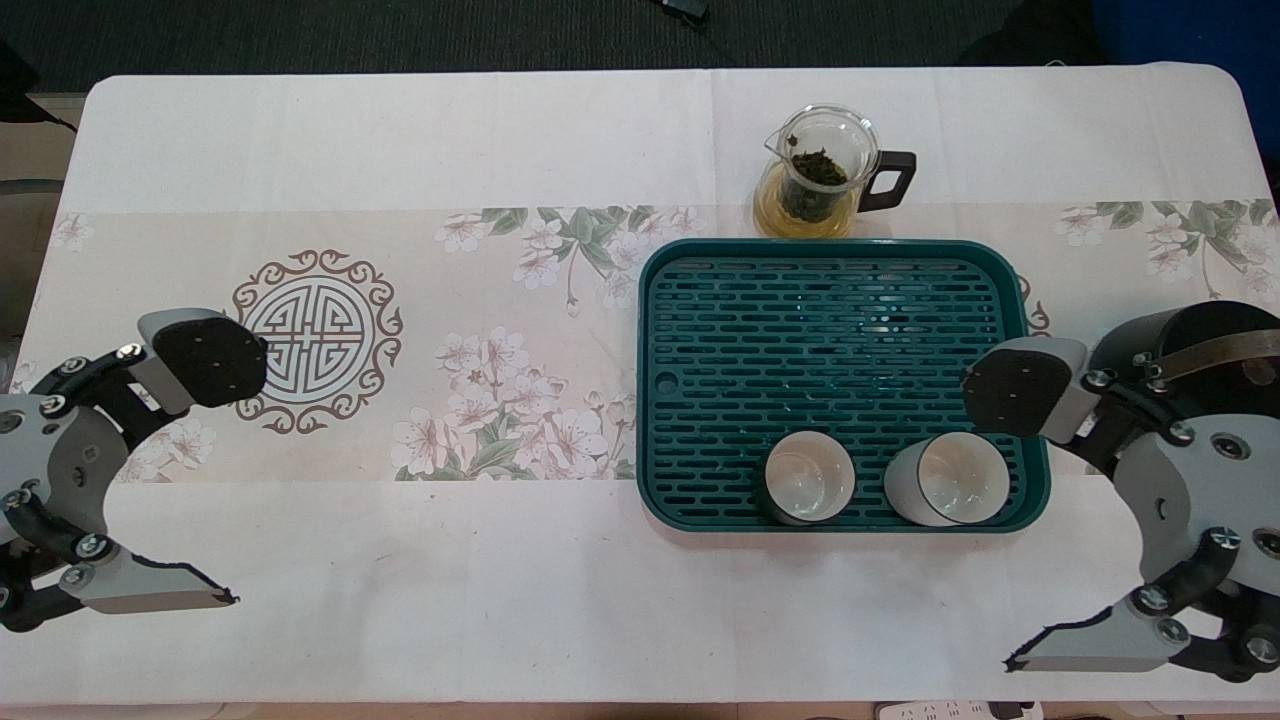}
    \caption{Pouring 4}
  \end{subfigure}\hfill
  \begin{subfigure}[t]{0.16\textwidth}
    \includegraphics[width=\linewidth]{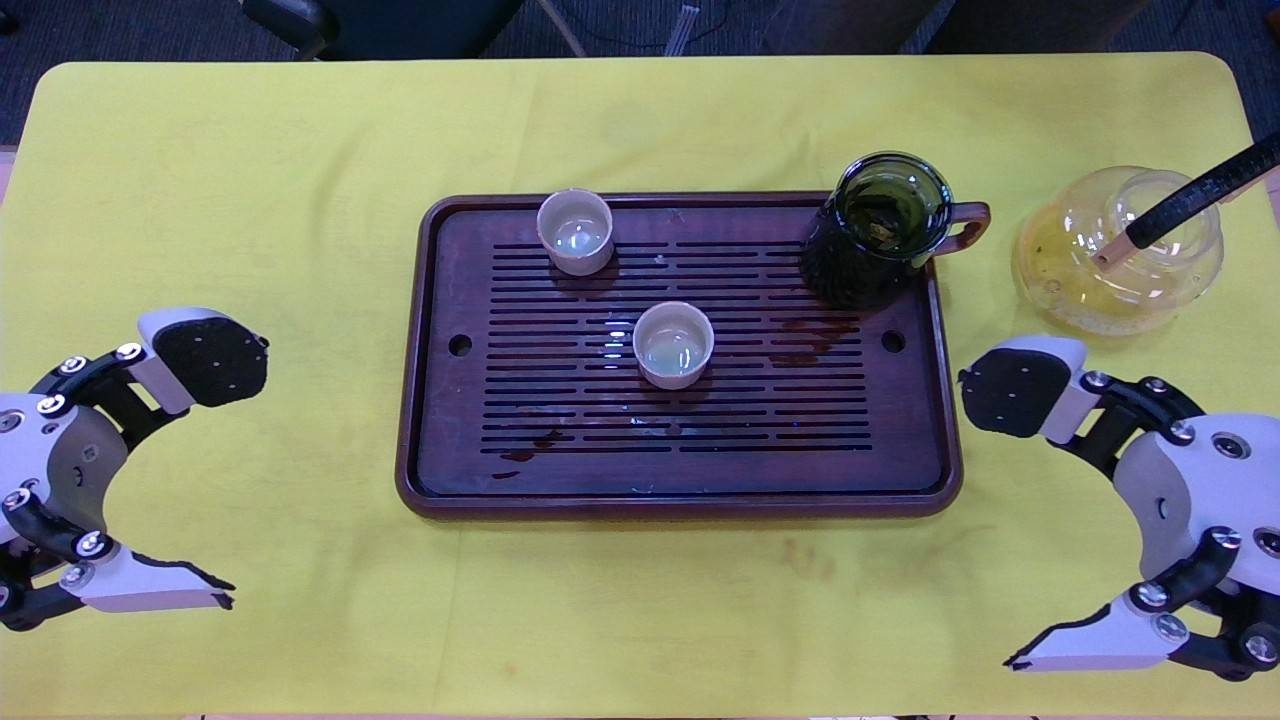}
    \caption{Pouring 5}
  \end{subfigure}\hfill
  \begin{subfigure}[t]{0.16\textwidth}
    \includegraphics[width=\linewidth]{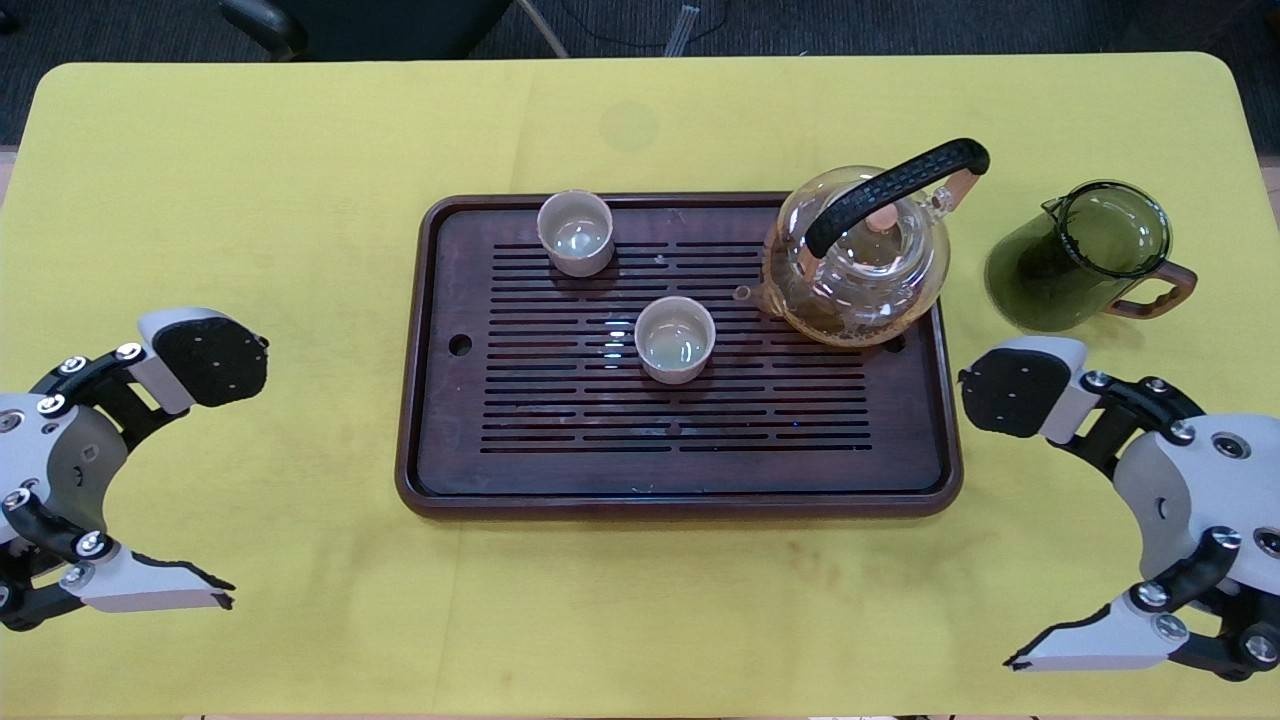}
    \caption{Pouring 6}
  \end{subfigure}
  \caption{\textbf{Representative evaluation placements on AGIBOT G1 tasks.} Top row (a--f): Pick--Place and Pick--Pass--Place across \emph{Seen} colored balls, \emph{Seen} cluttered table, and \emph{Unseen} scenes. Bottom row (g--l): Pour Water and Pour Tea.}
  \label{fig:eval_placements}
\end{figure}

At test time, we evaluate each task across \textbf{\emph{diverse initial object placements and instructions}} rather than testing on a single fixed scene (Fig.~\ref{fig:eval_placements}), using \textbf{\emph{the same set of placements and instructions for all models}}. Pick--Place and Pick--Pass--Place cover colored-ball (\emph{Seen}) and cluttered-table (\emph{Seen}) scenes, as well as \emph{Unseen} scenes with objects and tablecloths absent from the fine-tuning data, varying object positions and target instructions across trials. Note that \emph{Seen} denotes familiar objects rather than in-domain scenes: individual objects and tablecloths have appeared somewhere in the fine-tuning dataset, but the multi-object combinations, spatial layouts, and instructions at test time do not replicate training episodes, \textbf{making the \emph{Seen} setting an evaluation of compositional generalization}. Pour Water and Pour Tea are evaluated under \emph{Seen} across different vessel sets and placements, varying whether the kettle and teapot are placed on the tea tray or on the table, as well as the target cups and return locations.

\paragraph{Tasks with additional embodiments and dexterous hands.}
We further use an AGIBOT G2 robot and WUJI-1 dexterous hands mounted on Tianji Marvin arms to test our model. On AGIBOT G2, we design two tasks: \emph{Laundry unloading} with an OmniPicker gripper, where the robot removes clothes from a washing machine into a basket held by the other arm (Fig.~\ref{fig:real-rollouts}d), and \emph{Workpiece loading} with a DH gripper and custom fingertips, where the robot picks up workpieces at arbitrary poses, hands them over to re-orient them, and places them onto a conveyor belt (Fig.~\ref{fig:real-rollouts}e). The evaluations vary the initial scene conditions across trials, such as different clothing colors and poses inside the washing machine for Laundry unloading, and random part poses in the tray for Workpiece loading. For dexterous manipulation, we evaluate \emph{Bottle-cap unscrewing}, where after grasping a bottle, one hand stabilizes the bottle while the other unscrews the cap (Fig.~\ref{fig:real-rollouts}f), and \emph{Playing-card pickup}, where the index finger presses the card edge to tilt it up before pinching it with the thumb and placing it into a box (Fig.~\ref{fig:real-rollouts}g).

\paragraph{Fine-tuning data statistics.}
As noted in Sec.~\ref{sec:training}, we fine-tune all models on the exact same training data for each task under consistent experimental settings. Tables~\ref{tab:training_data} and~\ref{tab:adaptation_data} report the fine-tuning data statistics across all platforms. For the AGIBOT G1 dataset, beyond the three evaluated tasks, the data also include other skills (\emph{e.g.}, scooping with a spoon). The Pick/Pass/Place skills cover more than 100 objects, but not every object has training trajectories for every skill (\emph{e.g.}, bimanual handover), and Pour Water and Pour Tea cover multiple pouring vessels and target cups. During G1 data collection, table height, lighting, and tablecloths vary across episodes, while a program selects the objects, layouts, and initial arm poses and generates stage-level instructions automatically to avoid manual layout bias.

\subsubsection{Results and Analysis}
\label{sec:real}
\label{sec:real-results}
\label{app:real-robot}
\label{app:real-hardware-results}

\begin{table}[t!]
\centering\small\setlength{\tabcolsep}{4pt}
\caption{\textbf{AGIBOT G1 stage-wise and end-to-end success rates (\%).} Stage rates are conditional on reaching that stage, and their product gives the end-to-end (E2E) success rate. A dash indicates that no trial reached that stage.}
\label{tab:g1-stages}
\captionsetup[subtable]{position=top,font=small,labelfont=bf,justification=centering,singlelinecheck=false,skip=4pt}
\begin{subtable}{\linewidth}
\centering
\caption{Pick--Place}\label{tab:pp_results}
\begin{adjustbox}{max width=\linewidth}
\begin{tabular}{lrr>{\columncolor{reportaccent!8}}rrr>{\columncolor{reportaccent!8}}r}
    \toprule
     & \multicolumn{3}{c}{Seen} & \multicolumn{3}{c}{Unseen} \\
    \cmidrule(lr){2-4}\cmidrule(lr){5-7}
    Model & Pick & Place & E2E & Pick & Place & E2E \\
    \midrule
    $\pi_{0.5}$~\cite{black2025pi05}  & 40.0 & 83.3 & 33.3 & 20.0 & 100.0 & 20.0 \\
    StarVLA-$\pi$~\cite{starvla2025}  & 90.0 & 81.5 & 73.3 & 46.7 & 57.1 & 26.7 \\
    InternW0-$\Delta$~\cite{internw0} & 90.0 & 92.6 & 83.3 & 40.0 & 50.0 & 20.0 \\
    \midrule
    Ours--no pretrain & 83.3 & 80.0 & 66.7 & 46.7 & 100.0 & 46.7 \\
    \textbf{Ours} & 100.0 & 93.3 & \textbf{93.3} & 73.3 & 81.8 & \textbf{60.0} \\
    \bottomrule
  \end{tabular}
\end{adjustbox}
\end{subtable}
\par\medskip
\begin{subtable}{\linewidth}
\centering
\caption{Pick--Pass--Place}\label{tab:ppp_results}
\begin{adjustbox}{max width=\linewidth}
\begin{tabular}{lrrr>{\columncolor{reportaccent!8}}rrrr>{\columncolor{reportaccent!8}}r}
    \toprule
     & \multicolumn{4}{c}{Seen} & \multicolumn{4}{c}{Unseen} \\
    \cmidrule(lr){2-5}\cmidrule(lr){6-9}
    Model & Pick & Pass & Place & E2E & Pick & Pass & Place & E2E \\
    \midrule
    $\pi_{0.5}$~\cite{black2025pi05}  & 53.3 & 25.0 & 100.0 & 13.3 & 33.3 & 0.0 & --- & 0.0 \\
    StarVLA-$\pi$~\cite{starvla2025}  & 83.3 & 88.0 & 81.8 & 60.0 & 40.0 & 50.0 & 66.7 & 13.3 \\
    InternW0-$\Delta$~\cite{internw0} & 80.0 & 66.7 & 56.3 & 30.0 & 40.0 & 66.7 & 100.0 & 26.7 \\
    \midrule
    Ours--no pretrain & 86.7 & 96.2 & 80.0 & 66.7 & 60.0 & 44.4 & 100.0 & 26.7 \\
    \textbf{Ours} & 100.0 & 93.3 & 96.4 & \textbf{90.0} & 86.7 & 61.5 & 87.5 & \textbf{46.7} \\
    \bottomrule
  \end{tabular}
\end{adjustbox}
\end{subtable}
\par\medskip
\begin{subtable}{\linewidth}
\centering
\caption{Pour Water and Pour Tea}\label{tab:pour_results}
\begin{adjustbox}{max width=\linewidth}
\begin{tabular}{lrrr>{\columncolor{reportaccent!8}}rrrr>{\columncolor{reportaccent!8}}r}
    \toprule
     & \multicolumn{4}{c}{Pour Water} & \multicolumn{4}{c}{Pour Tea} \\
    \cmidrule(lr){2-5}\cmidrule(lr){6-9}
    Model & Pick & Pour & Place & E2E & Pick & Pour & Place & E2E \\
    \midrule
    $\pi_{0.5}$~\cite{black2025pi05}  & 73.3 & 27.3 & 66.7 & 13.3  & 20.0 & 33.3 & 100.0 & 6.7 \\
    StarVLA-$\pi$~\cite{starvla2025}  & 46.7 & 42.9 & 100.0 & 20.0 & 66.7 & 50.0 & 100.0 & 33.3 \\
    InternW0-$\Delta$~\cite{internw0} & 80.0 & 83.3 & 90.0 & 60.0 & 80.0 & 83.3 & 100.0 & 66.7 \\
    \midrule
    Ours--no pretrain & 80.0 & 91.7 & 90.9 & 66.7 & 80.0 & 91.7 & 100.0 & 73.3 \\
    \textbf{Ours} & 93.3 & 92.9 & 92.3 & \textbf{80.0} & 93.3 & 100.0 & 100.0 & \textbf{93.3} \\
    \bottomrule
  \end{tabular}
\end{adjustbox}
\end{subtable}
\par\medskip
\end{table}

\paragraph{AGIBOT G1 tasks.}
Table~\ref{tab:g1-stages}(a--c) reports the stage-wise conditional and end-to-end success rates on AGIBOT G1 alongside $\pi_{0.5}$~\cite{black2025pi05}, StarVLA-$\pi$~\cite{starvla2025}, and InternW0-$\Delta$~\cite{internw0}, all fine-tuned on the same data and evaluated under the same protocol. Our model achieves strong end-to-end performance and the highest success rates across Pick--Place (\emph{Seen} and \emph{Unseen}), Pick--Pass--Place (\emph{Seen} and \emph{Unseen}), Pour Water, and Pour Tea. Comparing our model with and without pretraining shows that action pretraining improves grasp quality and reliability (\emph{e.g.}, \emph{Unseen} Pick rises from 60.0\% to 86.7\% in Pick--Pass--Place, and handle grasping in both Pour Water and Pour Tea rises from 80.0\% to 93.3\%), stabilizes bimanual handover on unseen objects (\emph{Unseen} Pass from 44.4\% to 61.5\%), and enhances placement accuracy (\emph{Seen} Place from 80.0\% to 96.4\%).

\begin{table}[htbp]
\centering
\caption{\textbf{End-to-end success rates (\%) on AGIBOT G2 and WUJI-1 dexterous-hand tasks.} 
}
\label{tab:real}
\label{tab:g2_results}
\label{tab:dex_results}
\small
\setlength{\tabcolsep}{6pt}
\begin{adjustbox}{max width=\linewidth}
\begin{tabular}{lcccc}
\toprule
\multirow{2}{*}{Model} & \multicolumn{2}{c}{AGIBOT G2} & \multicolumn{2}{c}{WUJI-1 Hand} \\
\cmidrule(lr){2-3}\cmidrule(l){4-5}
& \shortstack{Laundry\\unloading} & \shortstack{Workpiece\\loading}
& \shortstack{Bottle-cap\\unscrewing} & \shortstack{Playing-card\\pickup} \\
\midrule
$\pi_{0.5}$~\cite{black2025pi05}  & ~~60.9 & 73.3 & ~~0.0 & ~~0.0 \\
InternW0-$\Delta$~\cite{internw0} & ~~82.6 & 83.3 & ~~0.0 & 26.7 \\
\midrule
Ours--no pretrain & \textbf{100.0} & 83.3 & ~~0.0 & ~~0.0 \\
\textbf{Ours} & \textbf{100.0} & \textbf{86.7} & \textbf{53.3} & \textbf{60.0} \\
\bottomrule
\end{tabular}
\end{adjustbox}
\end{table}

\paragraph{AGIBOT G2 tasks.}
Table~\ref{tab:g2_results} reports the end-to-end results on AGIBOT G2 and WUJI-1 dexterous-hand tasks. On AGIBOT G2, our model achieves strong performance with the highest success rates on both Laundry unloading (100.0\%) and Workpiece loading (86.7\%). Comparing our model with and without pretraining, both reach a 100.0\% success rate on Laundry unloading, though their ending behaviors differ: after all clothes are removed, the robot is expected to check inside the drum and retract its arm once empty. Our pretrained model retracts cleanly, while the model without pretraining stays near the drum opening without retracting. On Workpiece loading, pretraining improves end-to-end success from 83.3\% to 86.7\%.

\paragraph{Dexterous-hand tasks.}
On the two WUJI-1 dexterous-hand tasks (Table~\ref{tab:dex_results}), our model achieves 53.3\% and 60.0\% success on Bottle-cap unscrewing and Playing-card pickup, while our no-pretraining baseline achieves 0.0\% on both tasks. Because dexterous manipulation involves complex multi-finger motions and the fine-tuning dataset is relatively small (Table~\ref{tab:adaptation_data}), learning multi-finger control from fine-tuning alone is challenging, while pretraining with egocentric human-hand data enables our model to adapt effectively to dexterous hands.
\FloatBarrier
\begin{figure}[t!]
\centering
\includegraphics[width=1.0\linewidth]{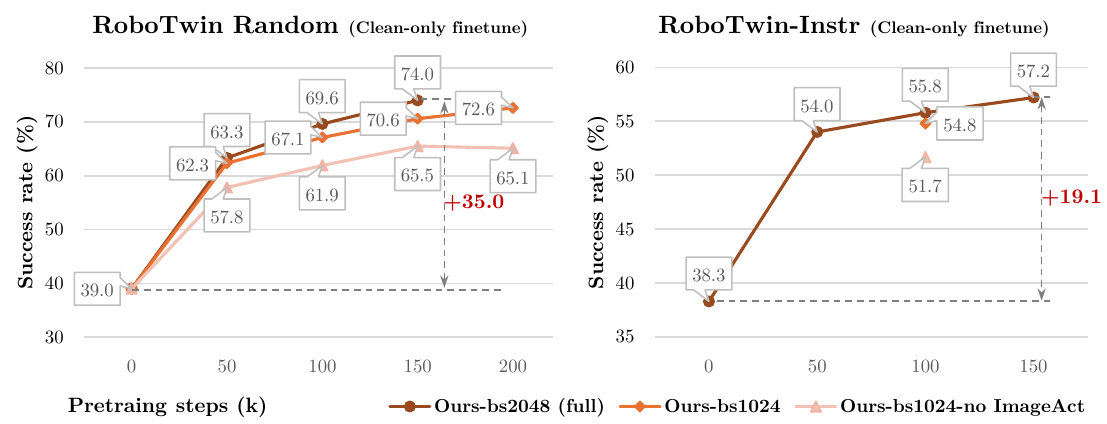}
\caption{
\textbf{Effect of model pretraining and ImageAct data supervision} on RoboTwin Random and RoboTwin-Instr, finetuned with Clean-only data. We compare the model variants pretrained with different batch sizes (bs) and with/without our ImageAct data.
}
\label{fig:pretraining-scaling}
\end{figure}

\section{Ablation Study and Analysis}
\label{sec:analysis}
\label{app:additional-analysis}
\label{app:ablation-settings}

\subsection{Effect of Pretraining}
Complementing the gains in real-robot task performance from pretraining (Sec.~\ref{app:real-robot}), \Cref{fig:pretraining-scaling} shows that pretraining also substantially improves model generalization performance on RoboTwin. Compared with no action pretraining, 150K pretraining steps with our full setting improve success rates by \emph{\textbf{35.0}} and \textbf{\emph{19.1}} percentage points for Clean-to-Random and RoboTwin-Instr, respectively. In-domain performance also improves, with Clean-set success increasing from 86.9\% to 90.4\% (+\emph{3.5} points).

\subsection{ImageAct Supervision}
\label{sec:imageact-analysis}

\paragraph{Effect of ImageAct supervision.}
We study the effect of the ImageAct data supervision, its sampling weight, advantage over VLM co-training, and backbone choices using a pretraining setup with a batch size of 1,024 and primarily at 100K steps. \Cref{tab:ablation-all}(a) shows that adding ImageAct data boosts success rates on RoboTwin clean-to-random transfer from 61.9\% to 63.0\%, RoboTwin-Instr from 51.7\% to 55.2\%, and RoboDojo from 13.1\% to 14.2\%. \Cref{fig:pretraining-scaling} shows that {\emph{the gains are stable or even increase with more pretraining steps}} on RoboTwin. These consistent gains demonstrate the value of image-derived action targets as complementary supervision to recorded trajectories. Adding auxiliary bounding-box supervision to the action transformer on ImageAct data further improves manipulation performance.

\begin{table}[!t]
\centering\small\setlength{\tabcolsep}{4pt}
\caption{\textbf{Supervision, sampling, and backbone ablations.} For more efficient pretraining and evaluation, all experiments in this table use a batch size of 1,024 and 100K pretraining steps; performance gaps are stable or even widen with longer pretraining, as suggested by \cref{fig:pretraining-scaling}.
RT denotes RoboTwin 2.0 with clean-only training. Panel (b) examines the effect of varying the robot:ImageAct:human sampling weights as $15{:}x{:}1$. The default sampling weights for Panel (a)(c)(d) are $15{:}5{:}1$, though our full pretraining uses $15{:}3{:}1$ (the two settings yield comparable overall performance as shown by the ablation study in Panel (b)).
}
\label{tab:ablation-all}
\begin{threeparttable}[b]
\begin{tabularx}{\linewidth}{@{}Xcccc@{}}
\toprule
Variant & RT-Clean & RT-Random & RT-Instr\tnote{$\dagger$} & RoboDojo \\
\midrule
\multicolumn{5}{l}{\textbf{(a) ImageAct supervision and action space}} \\
~~~~~~No ImageAct & 87.9 & 61.9 & 51.7 & 13.1 \\
~~~~~~+ action & 89.0 & 63.0 & \textbf{55.2} & 14.2 \\
~~~~~~+ action + box (\textbf{Ours}) & \textbf{89.5} & \textbf{64.4} & \textbf{55.2} & \textbf{15.0} \\
\midrule

~~~~~~Ours -- mixed space & 88.3 & 63.4 & 53.7 & 13.0 \\
\midrule
\multicolumn{5}{l}{\textbf{(b) ImageAct sampling weight $x$}} \\
~~~~~~0 (no ImageAct) & 87.9 & 61.9 & 51.7 & 13.1 \\
~~~~~~3 & 88.7 & \textbf{67.1} & 54.8 & 14.6 \\
~~~~~~5 & \textbf{89.5} & 64.4 & \textbf{55.2} & \textbf{15.0} \\
~~~~~~8 & 88.8 & 65.9 & 54.9 & 14.1 \\
\midrule
\multicolumn{5}{l}{\textbf{(c) ImageAct versus VLM co-training}} \\
~~~~~~ImageAct & \textbf{89.5} & \textbf{64.4} & \textbf{55.2} & \textbf{15.0} \\
~~~~~~VLM co-training & 85.1 & 59.2 & 41.3 & 12.9 \\
\midrule
\multicolumn{5}{l}{\textbf{(d) Pretrained VLM base model}} \\
~~~~~~RynnBrain1.1-2B & \textbf{89.5} & \textbf{64.4} & \textbf{55.2} & \textbf{15.0} \\
~~~~~~Qwen3.5-2B & 89.2 & 61.5 & 50.7 & 11.9 \\
\bottomrule
\end{tabularx}
    \begin{tablenotes}
      \footnotesize
      \item[$\dagger$] Note: For generalization evaluation on RoboTwin-Instr, we primarily use models trained on mixed Clean+Random data  (Tab.~\ref{tab:robotwin-if-main},\,\ref{tab:cfg}). For the ablation studies in this table as well as for Fig.~\ref{fig:pretraining-scaling}, however, we evaluate the model trained on Clean-only data  due to the high computational cost of mixed-data training.
    \end{tablenotes}
\end{threeparttable}
\end{table}

\paragraph{ImageAct data sampling weight.}We vary the pretraining sampling weights as robot:\allowbreak{}ImageAct:\allowbreak{}human $=15{:}x{:}1$, with $x\in\{0,3,5,8\}$, where $x=0$ excludes ImageAct data. As shown in \cref{tab:ablation-all}(b), including ImageAct with $x\in\{3,5,8\}$ consistently improves performance over $x=0$. The best results are achieved with either $x=3$ or $x=5$, depending on the benchmark, with comparable overall performance between the two settings. Increasing the weight to $x=8$ yields slightly weaker results.

%achieves the highest RT-Random success rate (67.1\%), while $x=5$ achieves the highest success rates on RT-Clean, RT-Instr, and RoboDojo (89.5\%, 55.2\%, and 15.0\%, respectively).

\paragraph{ImageAct versus VLM co-training.}We compare ImageAct with VLM co-training using the same image corpus. For the latter, we convert the available 2D and 3D bounding-box annotations into question--answer pairs following the grounding format of Qwen VLMs~\cite{bai2025qwen3,qwen3.5}, and train the VLM to predict boxes as text following the data sampling ratio of the ImageAct data. For VLM co-training, we weight the language-modeling loss by 0.1 relative to the action loss, following $\pi_{0.5}$~\cite{black2025pi05} and Qwen-VLA~\cite{wang2026qwenvla}.
Under this recipe, ImageAct outperforms VLM co-training by 4.4 and 5.2 percentage points on RT-Clean and RT-Random, respectively (\cref{tab:ablation-all}(c)).
These results suggest that ImageAct provides a more effective way to use geometric annotations for downstream manipulation.

\begin{table}[t!]
\centering
\caption{\textbf{Inference-time classifier-free guidance ablation.} Both CFG variants use a guidance scale of 1.5. In this table, RT-Clean and RT-Random results are evaluated with the Clean-only-training model, whereas RT-Instr evaluates the Clean+Random mixed-training model.}
\label{tab:cfg}
\footnotesize
\setlength{\tabcolsep}{1.2pt}

\begin{tabular*}{\columnwidth}{@{\extracolsep{\fill}}lccccc@{}}
\toprule
Guidance & RT-Clean & RT-Random & RT-Instr & LIBERO-Plus & LIBERO-Pro \\
\midrule
No CFG & 90.5 & 70.7 & 60.2 & 93.8 & 61.8 \\
V-L CFG & \textbf{90.9} & 70.9 & 60.2 & 94.0 & 62.1 \\
Language CFG\!\!\! & 90.4 & \textbf{74.0} & \textbf{61.2} & \textbf{94.6} & \textbf{63.2} \\
\bottomrule
\end{tabular*}
\end{table}

\subsection{Consistent Camera-Space Supervision}
We compare unified camera-space supervision with mixed-space training used in prior work~\cite{yuan2026qwenrobotmanip}. Specifically, we train a variant in which 21\% of the robot pretraining data (approximately 2K hours) uses robot-space rather than camera-space action targets. These data lack camera extrinsics in the original datasets, which our calibration pipeline (\cref{sec:camera-calibration}) recovers to enable camera-space supervision. \Cref{tab:ablation-all}(a) shows that mixed-space training lowers success rates from 64.4\% to 63.4\% on RoboTwin clean-to-random transfer, from 55.2\% to 53.7\% on RoboTwin-Instr, and from 15.0\% to 13.0\% on RoboDojo.

\subsection{Vision-Language Model}

In this experiment, we compare our VLM base model RynnBrain1.1-2B~\cite{li2026rynnbrain11} with the Qwen3.5-2B~\cite{qwen3.5} model and retain the same action transformer.
Both backbones achieve similar in-domain performance, with RT-Clean success rates of 89.5\% and 89.2\%, respectively. RynnBrain performs better on clean-to-random transfer, achieving 64.4\% on RT-Random compared with 61.5\% for Qwen3.5-2B, a gain of 2.9 percentage points (\cref{tab:ablation-all}(d)).

\subsection{Classifier-Free Guidance}
During training, we randomly drop language and visual inputs, enabling classifier-free guidance (CFG)~\cite{ho2022classifierfree} at inference. Although prior work has explored CFG for VLAs (\emph{e.g.},~\cite{pi_07}), its effects on generalization when guiding different modalities remain insufficiently understood.
We compare Language CFG, whose reference branch retains vision but removes language, with Vision--Language CFG, whose reference branch removes both modalities.
\Cref{tab:cfg} shows that Language CFG consistently improves generalization over No CFG across RoboTwin clean-to-random, RoboTwin-Instr, LIBERO-Plus, and LIBERO-Pro. Notably, it raises RoboTwin clean-to-random success from 70.7\% to 74.0\% (+\textbf{3.3} percentage points). These results demonstrate that strengthening language guidance improves robustness to scene and task perturbations beyond standard vision--language conditioning. In contrast, Vision--Language CFG yields only modest gains.

\section{Conclusion}
\label{sec:conclusion}
We have shown that a simple, compact VLA can achieve strong performance and generalization through effective supervision and pretraining. Our approach unifies robot demonstrations and introduces ImageAct to convert large-scale still-image geometry into action supervision. The resulting 2.4B model achieves strong results across multiple simulation benchmarks and three real-robot platforms. Ablations highlight the value of ImageAct, action-space consistency, auxiliary grounding, and language guidance, with pretraining yielding particularly large gains under distribution shift. Together, these findings suggest that model complexity is not the only path to capable robot policies: scalable data, aligned supervision, and effective pretraining can provide a strong foundation for simple, general-purpose robot control.

\subsection*{Acknowledgments}

We would like to thank PIA Automation, Business Segment Physical AI \& Humanoid Robotics for providing the training data and evaluation environments for experiments with the AGIBOT G2 robot, Wuji Technology for providing the training data for the dexterous manipulation experiments, Zhenyu Hu for help in real-world task evaluation, Yu Deng, Xu Liu, and Dong Chen for helpful discussions, and Lidong Zhou for strategic guidance on this project.
\FloatBarrier
\label{page:main-end}

\bibliography{references}
\bibliographystyle{plainnat}

\clearpage
\appendix
\renewcommand{\thesection}{\Alph{section}}
\renewcommand{\thesubsection}{\thesection.\arabic{subsection}}
\renewcommand{\theHsection}{appendix.\arabic{section}}
\renewcommand{\theHsubsection}{\theHsection.\arabic{subsection}}
\setcounter{equation}{0}
\setcounter{table}{0}
\setcounter{figure}{0}
\renewcommand{\theequation}{A\arabic{equation}}
\renewcommand{\thetable}{A\arabic{table}}
\renewcommand{\thefigure}{A\arabic{figure}}
\renewcommand{\theHequation}{appendix.\arabic{equation}}
\renewcommand{\theHtable}{appendix.\arabic{table}}
\renewcommand{\theHfigure}{appendix.\arabic{figure}}
\section*{Appendix}
\label{page:supplement-start}

\section{ImageAct Data Sample Generation Details}
\label{app:imageact}

\paragraph{Task and robot sampling.}

CA1M's 2D boxes are obtained by projecting 3D boxes and are looser than tight image-space boxes, so we omit auxiliary 2D bounding-box supervision for this dataset.

\paragraph{Initial-state sampling.}
For the \emph{arm-base poses}, the distance between the left and right bases ranges from
10 to 110\,cm.
We apply left--right symmetric random yaw perturbations of up to $2^\circ$ to the left and right arm bases relative to their respective fixed mounting orientations. We define a shared coordinate frame at the midpoint of the base reference points, with its axes pointing forward, left, and upward relative to the robot.

For the \emph{end-effector poses}, we construct a bank of
131K
representative poses for each robot based on clustering end-effector poses from the pretraining data. We sample end-effector poses from this bank, then add a translation perturbation sampled uniformly within a three-dimensional ball of radius 3\,cm and a rotation perturbation sampled uniformly within an $\mathrm{SO}(3)$ geodesic ball of radius $5^\circ$. We then use inverse kinematics to obtain the corresponding joint configurations.

The initial \emph{gripper openings} are sampled independently for the two grippers, with a 25\% probability each of being fully open or fully closed and the remaining 50\% sampled uniformly over the normalized gripper range $[0,1]$.

We sample the \emph{camera pose} relative to the shared robot frame and use the camera intrinsics associated with the geometrically augmented image. For tabletop arms, the lateral offset, fore--aft offset, and camera height are sampled independently and uniformly from base ranges of $[-5,5]$\,cm, $[-10,10]$\,cm, and $[45,75]$\,cm, respectively. We scale these translation ranges according to arm reach. Downward pitch and yaw are sampled from $[40^\circ,60^\circ]$ and $[-10^\circ,10^\circ]$, respectively. For rendering, we invert the sampled camera transform in the robot frame to obtain the robot pose in the image camera frame.

\paragraph{Action and instruction generation.}
We define task-specific key end-effector poses and interpolate between them to generate a trajectory. For tasks other than \emph{move to target pose}, we use a nominal translation speed of 0.4\,m/s and retain only the first action chunk for training. For \emph{move to target pose}, we interpolate the specified translation or rotation over a single action chunk.

We prepare at least 10 instruction templates for each task and fill them with the same parameters used to generate the actions. Image coordinates in the instructions are normalized to $[0,1]$. We show one example instruction per task below.

\begin{itemize}
\item \emph{Touch box center}: ``Touch the center of the mug. The action was synthetically generated.''
\item \emph{Touch surface point}: ``Touch the point (0.4200, 0.6100) on the mug. The action was synthetically generated.''
\item \emph{Move to target pose w.r.t. object}: ``Position your right gripper at the point halfway along the top-front edge of the box. Move in from overhead and keep the fingers lined up left to right, staying 2 cm off the surface. The action was synthetically generated.''
\item \emph{Move to target pose}: ``Use your left gripper to move 10 cm upward, opposite gravity. The action was synthetically generated.''
\item \emph{Pick up}: ``Approaching from the left side, reach for the mug and lift it with your left hand. Keep the gripper opening vertical. The action was synthetically generated.''
\item \emph{Push}: ``Approaching from the front, bring your right gripper to the box, close it, and push 10 cm toward the object's center. The action was synthetically generated.''
\item \emph{Rotate}: ``From above, use your right hand to grasp the box, then rotate it 45 degrees clockwise. The action was synthetically generated.''
\end{itemize}

\section{RoboTwin-Instr: Design and Evaluation}
\label{app:robotwin-if}

\paragraph{Benchmark design.}
RoboTwin-Instr evaluates whether manipulation policies correctly follow language instructions and generalize to diverse combinations of objects (including unseen objects), attributes, and task configurations. The benchmark design draws partial inspiration from \cite{yuan2026qwenrobotmanip}, while extending generalization evaluation across multiple dimensions, including object attributes, unseen objects, arm selection, and operation sequences. Built on RoboTwin 2.0~\cite{chen2025robotwin2}, it covers six dimensions of instruction following. Tasks reuse pressing, lifting, placing and stacking primitives represented in the original RoboTwin demonstrations, allowing us to test changed behavioral requirements using familiar manipulation operations.

A central design principle is to vary the instruction while keeping non-linguistic initial conditions fixed. Within each contrast group, episodes share pixel-identical initial RGB observations and identical robot states, with object identities, poses, and distractors held constant. 
The six suites probe different ways in which instructions specify the required behaviors:

\begin{itemize}
\item \textbf{Verb (action selection).}
The scene contains a stapler and two distractors, with instructions to either press or lift the stapler. The shared initial scene tests whether the policy selects the requested action under the same visual context.
\item \textbf{Noun (object-name grounding).}
The policy selects and lifts a named target among three distractors, testing instruction following on \emph{seen} and \emph{unseen} objects. Unseen objects come from RoboTwin's object pool but are absent from native-task demonstrations.
\item \textbf{Attribute (attribute-based selection).}
Each scene contains two objects distinguished by color (Red/Blue), decal (Cat/Dog), shape (Block/Bar), or size (Big/Small). Their target and distractor roles switch with the instruction.
\item \textbf{Arm (arm selection).}
The policy must lift a single block using the left or right arm, as instructed.
\item \textbf{Sequence (stacking order).}
The policy stacks red, green, and blue blocks in one of six bottom-to-top
orders specified by the instruction.
\item \textbf{Spatial (relative placement).}
The scene contains a movable target, a reference object, and one distractor. The policy must place the target left of, right of or on top of the reference.
\end{itemize}

Figure~\ref{fig:robotwin-if-tasks} illustrates these suites with representative tasks.

\paragraph{Evaluation protocol.}
The benchmark comprises 23 tasks across six suites. Each task has 20 expert-qualified blocks, with
one episode per task in each block, yielding 460 episodes in total for each policy. All policies use the same frozen task/seed pairs and identical recorded instructions from the unseen instruction-template split. We average success rates across modes within each dimension, then weight the six dimensions equally.

\begin{table}[t!]
    \centering
    \ReportTableStyle
    \caption{\textbf{LIBERO benchmark results.} Success rates (\%)
        on Spatial, Object, Goal, and Long suites, together with the  average. $^*$ denotes models without embodied pretraining; the best results are highlighted in bold.}
    \label{tab:libero}
    \newcommand{\LiberoMethodWidth}{4cm}
    \newcommand{\LiberoResultWidth}{1.4cm}

    \begin{adjustbox}{max width=\linewidth}
    \begin{tabular}{@{}>{\raggedright\arraybackslash}p{\LiberoMethodWidth}l*{4}{>{\raggedright\arraybackslash}p{\LiberoResultWidth}}>{\columncolor{reportaccent!8}\raggedright\arraybackslash}p{\LiberoResultWidth}@{}}
        \toprule
        Model & Size (B) & Spatial & \!\!\!Object & Goal & Long & \textbf{Average} \\
        \midrule
            $\pi_{0.5}$~\cite{black2025pi05} & 3.4 & 98.8 & 98.2 & 98.0 & 92.4 & 96.9 \\
            X-VLA~\cite{zheng2025xvla} & 0.9 & 98.2 & 98.6 & 97.8 & 97.6 & 98.1 \\
            Motus~\cite{bi2026motus} & 8.2 & 96.8 & 99.8 & 96.6 & 97.6 & 97.7 \\
            Cosmos Policy$^*$~\cite{kim2026cosmospolicy} & 2.0 & 98.1 & \!\!\!\textbf{100.0} & 98.2 & 97.6 & 98.5 \\
            LingBot-VA~\cite{li2026lingbotva} & 5.5 & 98.5 & 99.6 & 97.2 & 98.5 & 98.5 \\
            VLA-JEPA~\cite{sun2026vlajepa} & 2.3 & 96.2 & 99.6 & 97.2 & 95.8 & 97.2 \\
            ABot-M0~\cite{yang2026abotm0} & 5.6 & 98.8 & 99.8 & 99.0 & 96.6 & 98.6 \\
            SimVLA$^*$~\cite{luo2026simvla} & 0.5 & \textbf{99.6} & 99.8 & 98.6 & 96.4 & 98.6 \\
            Fast-WAM$^*$~\cite{yuan2026fastwam} & 6.2 & 98.2 & \!\!\!\textbf{100.0} & 97.0 & 95.2 & 97.6 \\
            GR00T-N1.7~\cite{nvidia2026grootn17} & 3.1 & 97.7 & 98.5 & 97.5 & 94.4 & 97.0 \\
            MolmoAct2~\cite{fang2026molmoact2} & 5.0 & 97.8 & \!\!\!\textbf{100.0} & 97.8 & 93.2 & 97.2 \\
            Qwen-RobotManip~\cite{yuan2026qwenrobotmanip} & 4.9 & --- & --- & --- & --- & 99.1 \\
            ImageWAM$^*$~\cite{zhang2026imagewam} & 4.5 & 97.2 & 99.2 & 98.8 & 98.4 & 98.4 \\
            4D-WAM$^*$~\cite{yang2026fourdwam} & 6.2 & 98.8 & 99.5 & 98.6 & 97.4 & 98.6 \\
            Flex-$\pi$~\cite{yan2026flexpi} & 6.3 & \textbf{99.6} & 99.8 & 98.6 & 96.0 & 98.5 \\
            OpenWAM-$\alpha$~\cite{wang2026openwam} & 6.2 & \textbf{99.6} & 99.6 & \textbf{99.8} & 98.2 & 99.3 \\
            InternW0~\cite{cai2026internw0} & 7.6 & 99.4 & 99.4 & 98.6 & 97.0 & 98.6 \\
            UniWAM~\cite{song2026uniwam} & 8.2 & \textbf{99.6} & 99.6 & 99.2 & 98.4 & 99.2 \\
            \midrule
            \rowcolor{reportaccent!8}
            \textbf{Ours} & 2.4 & 99.2{\tiny $\pm$0.28} & \!\!\!\textbf{100.0}{\tiny $\pm$0.09} & 99.7{\tiny $\pm$0.23} & \textbf{98.7}{\tiny $\pm$0.44} & \textbf{99.4}{\tiny \,$\pm$0.15} \\
            \bottomrule
    \end{tabular}
    \end{adjustbox}
\end{table}

\section{Simulation Benchmark Protocols and Detailed Results}
\label{app:simulation-results}
\label{app:libero-results}

\subsection{Simulation Benchmark Protocols}
\label{app:simulation}
\label{app:simulation-protocols}
\label{app:robodojo}

\paragraph{LIBERO.}
The standard LIBERO~\cite{liu2023libero} evaluation uses four suites: Spatial, Object, Goal, and Long (also called LIBERO-10), covering spatial relationships, object identities, task goals, and long-horizon manipulation, respectively. Each suite contains 10 tasks with 50 human-teleoperated demonstrations per task, totaling 2,000 demonstrations across 40 tasks.

\paragraph{LIBERO-Plus.}
LIBERO-Plus~\cite{liberoplus2025} extends these 40 tasks to 10,030 evaluation instances to assess robustness under controlled distribution shifts. It preserves task semantics while varying seven dimensions: camera viewpoints, robot initial states, language instructions, lighting, background textures, sensor noise, and object layouts. Under the zero-shot robustness protocol, policies are adapted using standard LIBERO demonstrations and evaluated on LIBERO-Plus without further adaptation to the perturbed environments.

\paragraph{LIBERO-Pro.}
We evaluate four dimensions of LIBERO-Pro~\cite{zhou2025liberopro}: Object changes object appearance, color, and scale; Position relocates objects within feasible spatial bounds; Semantic paraphrases instructions while preserving their meaning; and Task changes target objects or goal conditions to test whether policies respond to the requested task rather than replay memorized behaviors. We report success rates for each dimension and their unweighted average.

\paragraph{RoboTwin 2.0.}
RoboTwin 2.0~\cite{chen2025robotwin2} contains 50 tasks, including object placement, stacking, handovers, and articulated-object manipulation. The framework generates expert demonstrations and supports domain randomization across five dimensions: distractor objects, background textures, lighting conditions, tabletop height, and language instructions.

\begin{table}[t!]
    \centering
    \ReportTableStyle
    \caption{\textbf{LIBERO-Plus benchmark results.} Zero-shot transfer from LIBERO-finetuned models. Success rates (\%) on Camera, Robot, Language, Light, Background, Noise, and Layout variation settings, together with the total average. $^*$ denotes models without embodied pretraining; the best results are highlighted in bold.}
    \label{tab:liberoplus}
    \setlength{\tabcolsep}{4pt}

    \begin{adjustbox}{max width=\linewidth}
    \begin{tabular}{@{}ll*{7}{l}>{\columncolor{reportaccent!8}}l@{}}
        \toprule
        Model & \shortstack{Size (B)} & Cam. & Robot & Lang. & Light & Bg. & Noise & Layout & \textbf{Total} \\
        \midrule
            $\pi_{0.5}$~\cite{black2025pi05} & 3.4 & 78.4 & 73.6 & 80.8 & 96.2 & 94.1 & 89.0 & 84.5 & 84.4 \\
            Cosmos Policy$^*$~\cite{kim2026cosmospolicy} & 2.0 & 75.8 & 63.3 & 81.7 & 96.5 & 88.9 & 92.7 & 82.2 & 82.2 \\
            VLA-JEPA~\cite{sun2026vlajepa} & 2.3 & 63.3 & 67.1 & 85.4 & 95.6 & 93.6 & 66.3 & 85.1 & 79.5 \\
            ABot-M0~\cite{yang2026abotm0} & 5.6 & 60.4 & 67.9 & 86.4 & 96.2 & 91.6 & 86.4 & 82.6 & 80.5 \\
            Qwen-RobotManip~\cite{yuan2026qwenrobotmanip} & 4.9 & 87.2 & 75.5 & 85.6 & 96.6 & 97.7 & 97.7 & 87.3 & 89.0 \\
            ImageWAM$^*$~\cite{zhang2026imagewam} & 4.5 & 80.8 & 50.3 & 91.4 & 98.1 & 85.5 & 93.8 & 80.5 & 83.1 \\
            InternVLA-A1.5~\cite{internvla_a15} & 2.7 & 83.1 & 55.1 & 86.9 & 96.4 & 98.2 & 95.6 & 85.2 & 84.8 \\
            4D-WAM$^*$~\cite{yang2026fourdwam} & 6.2 & 45.2 & 64.3 & 90.6 & 94.3 & 57.7 & 69.1 & 79.2 & 71.0 \\
            Flex-$\pi$~\cite{yan2026flexpi} & 6.3 & 89.1 & 72.7 & 84.5 & 94.6 & 96.3 & \textbf{98.9} & 87.3 & 88.6 \\
            VLAct~\cite{vlact2026} & 4.5 & 73.9 & 68.4 & 81.5 & 96.7 & 96.7 & 86.0 & 83.3 & 82.6 \\
            OpenWAM-$\alpha$~\cite{wang2026openwam} & 6.2 & 33.8 & 76.1 & 88.0 & 97.0 & 87.1 & 39.8 & 77.5 & 69.2 \\
            InternW0-$\Delta$~\cite{internw0} & 8.6 & 90.6 & 91.1 & 92.9 & 95.8 & \textbf{99.4} & 94.0 & 88.4 & 92.8 \\
            UniWAM~\cite{song2026uniwam} & 8.2 &  \textbf{92.0} &  89.5 &  92.2 &  97.9 &  97.6 & 94.3 &  84.6 & 92.6 \\
            \midrule
            \rowcolor{reportaccent!8}
            \textbf{Ours} & 2.4 & 91.1{\tiny $\pm$0.69} & \textbf{93.0}{\tiny $\pm$0.35} & \textbf{95.5}{\tiny $\pm$0.46} & \textbf{99.6}{\tiny $\pm$0.26} & \textbf{99.4}{\tiny $\pm$0.32} & 95.9{\tiny $\pm$0.36} & \textbf{90.2}{\tiny $\pm$0.17} & \textbf{94.6}{\tiny \,$\pm$0.24} \\
            \bottomrule
    \end{tabular}
    \end{adjustbox}
\end{table}

\paragraph{RoboDojo.}
The benchmark~\cite{chen2026robodojo} covers 42 simulation tasks spanning
Generalization, Precision, Long-Horizon, Memory, and Open. Its success-rate
aggregate averages tasks within each category and then weights the five
category means equally.

\subsection{Detailed Results}
Tables~\ref{tab:libero}, \ref{tab:liberoplus}, and~\ref{tab:liberopro} report success rates of different models on the LIBERO suites, LIBERO-Plus perturbations, and LIBERO-Pro evaluation dimensions, respectively.

Table~\ref{tab:robotwin-task-results} reports the per-task success rates of our model on RoboTwin 2.0 under the clean-only and clean+random training settings.

\begin{table}[t!]
    \centering
    \ReportTableStyle
    \caption{\textbf{LIBERO-Pro benchmark results.} Zero-shot transfer from LIBERO-finetuned models. Success rates (\%) on Object, Position, Semantic, and Task evaluations, together with their average. Our results report the mean and standard deviation over five runs. $^*$ denotes models without embodied pretraining; the best results are highlighted in bold.
    }
    \label{tab:liberopro}

    \begin{adjustbox}{max width=\linewidth}
    \begin{tabular}{@{}ll*{4}{l}>{\columncolor{reportaccent!8}}l@{}}
        \toprule
        Model & Size (B) & Object & Position & Semantic & Task & \textbf{Average} \\
        \midrule
            $\pi_{0.5}$~\cite{black2025pi05} & 3.4 & 85.4 & 28.2& 95.5 & 25.3 & 58.6 \\
            X-VLA~\cite{zheng2025xvla} & 0.9 & 79.0 & \ \,0.8 & 96.8 & \ \,6.8 & 45.8 \\
            Spatial Forcing~\cite{li2026spatial} & 7.7 & 74.0 & \ \,2.1 & 94.5 & \ \,9.8 & 45.1 \\
            Cosmos Policy$^*$~\cite{kim2026cosmospolicy} & 2.0 & 83.3 & \ \,0.9 & 94.8 & 22.9 & 50.5 \\
            VLA-JEPA~\cite{sun2026vlajepa} & 2.3 & 80.2 & \ \,4.2 & 97.4 & 22.1 & 51.0 \\
            SimVLA$^*$~\cite{luo2026simvla} & 0.5 & 81.5 & \ \,8.3 & 98.8 & \ \,6.0 & 48.6 \\
            Fast-WAM$^*$~\cite{yuan2026fastwam} & 6.2 & 71.5 & \ \,0.2 & 80.3 & 14.4 & 41.6 \\
            InternVLA-A1.5~\cite{internvla_a15} & 2.7 & 84.1 & 11.0 & 97.6 & 29.0 & 55.4 \\
            4D-WAM$^*$~\cite{yang2026fourdwam} & 6.2 & 79.9 & \ \,3.1 & 95.6 & 20.3 & 49.7 \\
            Flex-$\pi$~\cite{yan2026flexpi} & 6.3 & 87.3 & 11.8 & 98.3 & \ \,5.0 & 50.6 \\
            OpenWAM-$\alpha$~\cite{wang2026openwam} & 6.2 & 82.2 & 23.7 & 95.8 & 23.6 & 56.3 \\
            \midrule
            \rowcolor{reportaccent!8}
            \textbf{Ours} & 2.4 & \textbf{87.7}{\tiny \,$\pm$0.5} & \textbf{32.7}{\tiny \,$\pm$0.3} & \textbf{98.9}{\tiny \,$\pm$0.2} & \textbf{33.5}{\tiny \,$\pm$0.3} & \textbf{63.2}{\tiny \,$\pm$0.2} \\
            \bottomrule
    \end{tabular}
    \end{adjustbox}
\end{table}

\begin{table}[tp]
\centering
\footnotesize
\setlength{\tabcolsep}{6pt}
\renewcommand{\arraystretch}{1.1}
\caption{\textbf{Per-task results of our method on RoboTwin~2.0.} Success rates (\%) for the clean-only and clean+random training settings, evaluated on 100 episodes per task under each scene condition. Average is the mean of Clean and Random; the final row averages all 50 tasks.}
\label{tab:robotwin-task-results}

\begin{tabularx}{\textwidth}{@{}l*{6}{>{\raggedleft\arraybackslash}X}@{}}
\toprule
& \multicolumn{3}{c}{Clean-only training} & \multicolumn{3}{c}{Clean+Random training} \\
\cmidrule(lr){2-4}\cmidrule(l){5-7}
Task & Clean & Random & Average & Clean & Random & Average \\
\midrule
Adjust Bottle & 100.0 & 87.0 & 93.5 & 100.0 & 98.0 & 99.0 \\
Beat Block Hammer & 99.0 & 54.0 & 76.5 & 100.0 & 93.0 & 96.5 \\
Blocks Ranking RGB & 100.0 & 98.0 & 99.0 & 100.0 & 100.0 & 100.0 \\
Blocks Ranking Size & 77.0 & 68.0 & 72.5 & 95.0 & 88.0 & 91.5 \\
Click Alarmclock & 99.0 & 94.0 & 96.5 & 99.0 & 100.0 & 99.5 \\
Click Bell & 100.0 & 99.0 & 99.5 & 100.0 & 100.0 & 100.0 \\
Dump Bin Bigbin & 96.0 & 44.0 & 70.0 & 99.0 & 93.0 & 96.0 \\
Grab Roller & 100.0 & 100.0 & 100.0 & 100.0 & 100.0 & 100.0 \\
Handover Block & 94.0 & 65.0 & 79.5 & 97.0 & 91.0 & 94.0 \\
Handover Mic & 99.0 & 93.0 & 96.0 & 95.0 & 99.0 & 97.0 \\
Hanging Mug & 39.0 & 43.0 & 41.0 & 69.0 & 61.0 & 65.0 \\
Lift Pot & 100.0 & 43.0 & 71.5 & 100.0 & 100.0 & 100.0 \\
Move Can Pot & 96.0 & 42.0 & 69.0 & 99.0 & 99.0 & 99.0 \\
Move Pillbottle Pad & 99.0 & 43.0 & 71.0 & 100.0 & 97.0 & 98.5 \\
Move Playingcard Away & 99.0 & 98.0 & 98.5 & 100.0 & 100.0 & 100.0 \\
Move Stapler Pad & 67.0 & 57.0 & 62.0 & 97.0 & 91.0 & 94.0 \\
Open Laptop & 97.0 & 82.0 & 89.5 & 98.0 & 100.0 & 99.0 \\
Open Microwave & 99.0 & 61.0 & 80.0 & 100.0 & 100.0 & 100.0 \\
Pick Diverse Bottles & 85.0 & 65.0 & 75.0 & 88.0 & 94.0 & 91.0 \\
Pick Dual Bottles & 98.0 & 86.0 & 92.0 & 97.0 & 99.0 & 98.0 \\
Place A2B Left & 91.0 & 87.0 & 89.0 & 99.0 & 100.0 & 99.5 \\
Place A2B Right & 92.0 & 81.0 & 86.5 & 98.0 & 97.0 & 97.5 \\
Place Bread Basket & 93.0 & 82.0 & 87.5 & 89.0 & 92.0 & 90.5 \\
Place Bread Skillet & 84.0 & 70.0 & 77.0 & 89.0 & 93.0 & 91.0 \\
Place Burger Fries & 97.0 & 98.0 & 97.5 & 100.0 & 98.0 & 99.0 \\
Place Can Basket & 84.0 & 47.0 & 65.5 & 84.0 & 77.0 & 80.5 \\
Place Cans Plasticbox & 99.0 & 90.0 & 94.5 & 97.0 & 97.0 & 97.0 \\
Place Container Plate & 99.0 & 87.0 & 93.0 & 99.0 & 98.0 & 98.5 \\
Place Dual Shoes & 90.0 & 61.0 & 75.5 & 99.0 & 95.0 & 97.0 \\
Place Empty Cup & 100.0 & 90.0 & 95.0 & 100.0 & 100.0 & 100.0 \\
Place Fan & 93.0 & 89.0 & 91.0 & 97.0 & 93.0 & 95.0 \\
Place Mouse Pad & 73.0 & 57.0 & 65.0 & 99.0 & 97.0 & 98.0 \\
Place Object Basket & 81.0 & 70.0 & 75.5 & 87.0 & 94.0 & 90.5 \\
Place Object Scale & 97.0 & 75.0 & 86.0 & 97.0 & 95.0 & 96.0 \\
Place Object Stand & 99.0 & 88.0 & 93.5 & 97.0 & 95.0 & 96.0 \\
Place Phone Stand & 88.0 & 87.0 & 87.5 & 97.0 & 99.0 & 98.0 \\
Place Shoe & 99.0 & 90.0 & 94.5 & 100.0 & 98.0 & 99.0 \\
Press Stapler & 88.0 & 83.0 & 85.5 & 99.0 & 98.0 & 98.5 \\
Put Bottles Dustbin & 89.0 & 42.0 & 65.5 & 91.0 & 76.0 & 83.5 \\
Put Object Cabinet & 46.0 & 35.0 & 40.5 & 90.0 & 91.0 & 90.5 \\
Rotate QRcode & 88.0 & 64.0 & 76.0 & 93.0 & 99.0 & 96.0 \\
Scan Object & 90.0 & 64.0 & 77.0 & 99.0 & 95.0 & 97.0 \\
Shake Bottle Horizontally & 100.0 & 100.0 & 100.0 & 100.0 & 99.0 & 99.5 \\
Shake Bottle & 100.0 & 98.0 & 99.0 & 100.0 & 100.0 & 100.0 \\
Stack Blocks Three & 86.0 & 71.0 & 78.5 & 93.0 & 92.0 & 92.5 \\
Stack Blocks Two & 100.0 & 98.0 & 99.0 & 100.0 & 100.0 & 100.0 \\
Stack Bowls Three & 89.0 & 72.0 & 80.5 & 88.0 & 85.0 & 86.5 \\
Stack Bowls Two & 95.0 & 88.0 & 91.5 & 99.0 & 95.0 & 97.0 \\
Stamp Seal & 70.0 & 51.0 & 60.5 & 95.0 & 99.0 & 97.0 \\
Turn Switch & 76.0 & 61.0 & 68.5 & 87.0 & 88.0 & 87.5 \\
\midrule
\rowcolor{reportaccent!8}
\textbf{Average} & \textbf{90.4} & \textbf{74.0} & \textbf{82.2} & \textbf{95.9} & \textbf{94.8} & \textbf{95.3} \\
\bottomrule
\end{tabularx}
\end{table}

\FloatBarrier
\label{page:supplement-content-end}
\label{page:supplement-end}
\end{document}